\documentclass[preprint]{vgtc}               %

\graphicspath{{figures/}{pictures/}{images/}{./}} %

\usepackage{times}                     %

\usepackage{tabu}                      %
\usepackage{booktabs}                  %
\usepackage{lipsum}                    %
\usepackage{mwe}                       %

\usepackage{cite}
\usepackage{etoolbox}
\usepackage[table,svgnames]{xcolor}
\usepackage{pifont}

\definecolor{yellow}{RGB}{251, 210, 106}
\definecolor{mygreen}{RGB}{170, 215, 160}
\definecolor{myred}{RGB}{230, 150, 140}
\usepackage{amsmath}
\newcommand{\etal}{\textit{et al.}}
\newcommand{\supptitle}[1]{\def\mySuppTitlecontent{#1}}
\newcommand{\makeMySupplementaryTitle}{%
\clearpage
\onecolumn
\begin{center}
\vspace*{1em}
{\LARGE\bfseries \mySuppTitlecontent\par}
\vspace{1.5em}
{\large\itshape\textemdash\quad Supplementary Materials \quad\textemdash \par}
\vspace{2em}
\end{center}
}
\definecolor{green}{RGB}{152, 194, 130}
\definecolor{red}{RGB}{230, 117, 101} 
\definecolor{red_darker}{RGB}{204, 64, 64}
\definecolor{green_darker}{RGB}{112, 154, 90}

\newcommand{\bggreen}{\cellcolor{green}}  
\newcommand{\bgred}{\cellcolor{red}} 
\newcommand{\bgyellow}{\cellcolor{yellow}}
\newcommand{\fillcolor}[2]{
  \begingroup
  \setlength{\fboxsep}{1pt}
  \colorbox{#1!80}{\strut #2}
  \endgroup
}

\title{VAComp: View-Aware Full Projector Compensation}

\author{\texorpdfstring{\authororcid{Kai Zhou\thanks{e-mail: a13310885059@email.swu.edu.cn}}{0009-0003-8379-292X}%
\and \authororcid{Qingyue Deng\thanks{e-mail: dengqingyue@westlake.edu.cn}}{0009-0001-8471-2480}%
\and \setcounter{footnote}{3}\authororcid{Bingyao Huang\thanks{e-mail: bhuang@swu.edu.cn. Corresponding author.}}{0000-0002-8647-5730}}{Kai Zhou, Qingyue Deng, Bingyao Huang}}
\affiliation{\scriptsize
$^{\ast,\S}$College of Computer and Information Science, Southwest University, Chongqing, China\\
$^{\dagger}$School of Engineering, Westlake University, Hangzhou, Zhejiang, China}
\preprinttext{}
\nocopyrightspace

\teaser{
  \centering
  \includegraphics[width=\textwidth]{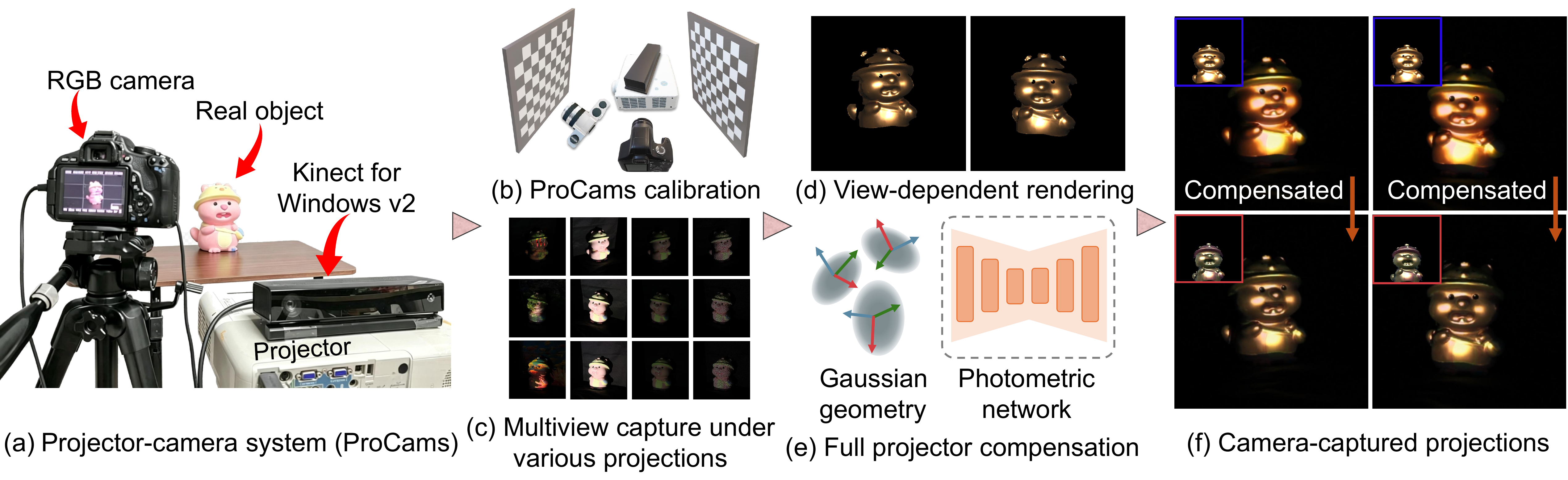}
  \caption{\textbf{VAComp: View-Aware Full Projector Compensation.} (a) Our ProCams setup consists of a projector, an RGB camera, and a Kinect for Windows v2 (hereafter Kinect v2), which tracks the viewer's eye pose. (b) We first establish a unified coordinate system using the initial pose of the RGB camera as the reference frame and calibrate the projector and Kinect v2 accordingly. (c) After calibration, we perform multiview capture under various projection patterns for training. (d) At runtime, the system dynamically renders the virtual object from the viewer’s eye pose. (e) VAComp combines Gaussian-based geometry and rendering with setup-dependent or setup-independent photometric compensation, resulting in VAComp-SD and VAComp-SI. (f) Camera-captured projection. The first row shows the camera-captured uncompensated projections from two different viewpoints. The images inside the blue boxes show the virtual object rendered from the projector viewpoint; directly projecting the images onto the surface results in the uncompensated projections captured by the camera at the corresponding viewpoints. The second row shows the corresponding camera-captured compensated projections generated by VAComp-SD from the same viewpoints. The images inside the red boxes show the compensation images at the corresponding viewpoints. For better visualization, the image shown in (d) and the camera-captured projections shown in (f) have been cropped and enlarged.
  }
  \label{fig:teaser}
}

\begin{document}

\abstract{
Full projector compensation adjusts a projector’s input image to compensate for both geometric and photometric distortions introduced by the projection surface. Existing methods generally treat the viewing condition as a static camera viewpoint. When applied to a moving viewer, existing methods face two key challenges: (1) view-dependent methods require model retraining or recalibration to compensate for novel viewpoints; and (2) NeRF/GS-based view-agnostic methods use structure from motion (SfM) to calibrate the projector-camera systems (ProCams) with an unknown global scale, while view-aware tracking requires calibrating a camera oriented opposite to the projector, which introduces an additional calibration challenge. To address these challenges, we introduce \textbf{VAComp}, the first \textbf{V}iew-\textbf{A}ware framework for full projector \textbf{Comp}ensation for a moving viewer in a naked-eye setting. Using a projector, VAComp enables the viewer to perceive a virtual object that visually replaces the real object or to observe view-consistent modifications of the real object’s appearance and material properties. To resolve the SfM scale ambiguity, we use metric calibration as a reference to recover the global scale; to calibrate the oppositely facing camera, we use two checkerboards, one facing the RGB camera and the other facing the depth camera. To balance compensation quality and runtime efficiency, we develop two non-iterative hybrid methods, VAComp-SD and VAComp-SI. Both combine 2D Gaussian-based view-conditioned geometric correction with learned photometric compensation, predicting each compensation image through a single forward pass without per-frame iterative optimization. Real-world experiments show that both methods achieve competitive compensation quality, with VAComp-SD enabling real-time updates and VAComp-SI operating at interactive rates. Project page: \url{https://realkaizhou.github.io/VAComp/}.
}
\keywords{Projector compensation, projection mapping, geometric calibration}

\firstsection{Introduction}
\maketitle

\label{sec:introduction}

Spatial Augmented Reality (SAR) produces the visual effect of seamlessly integrating virtual content with the physical environment, typically achieved through Projection Mapping (PM)~\cite{KhuslenBattulga2021,raskar2001shader,Lee2015Real,Siegl2015Real,Fujimoto2014Geometrically,deng2025,Meschini2016Expanded,Raskar1999Spatially,Erel2023Neural,Bluteau2005VisualSF,Chien2022APA,Huang2021DeProCams}. A projector displays virtual content onto projection surfaces to create a perceptual fusion between virtual content and the real environment, making the virtual objects appear as if they truly exist within the real world. To enhance the visual realism of 3D virtual objects, a view-dependent approach ensures that a virtual object's appearance remains correct by rendering it according to the viewer's viewpoint~\cite{Raskar1998EfficientIG,Arthur1993Evaluating,Bimber2005Enabling,KhuslenBattulga2021,Raskar1999Spatially}. However, when the surface has complex geometry and texture, directly projecting images can lead to undesired visual artifacts. Full projector compensation~\cite{Huang2021Full,Wang2023CompenHREF,Li2026SetupIndependentFP,Li2024physical,deng2025,Li2025DPCS,Asayama2018FabricatingDV,Bimber2005Embedded,Harville2006Practical} addresses this problem by altering the projector input image to compensate for the undesirable effects observed by a camera from specific viewpoints.

Although considerable effort has been devoted to developing full projector compensation methods, most existing  methods generally assume the viewing condition as a static camera viewpoint, making their extension to a moving viewer challenging. View-dependent methods~\cite{Huang2021Full,Wang2023CompenHREF,Li2026SetupIndependentFP} require model retraining or recalibration to compensate for novel viewpoints, interrupting the viewer's experience as the viewpoint changes. View-agnostic methods~\cite{Erel2023Neural,deng2025} support compensation from novel viewpoints without model retraining, but they still require calibration for each new viewpoint, increasing the workload. Moreover, viewers must move to the calibrated camera viewpoint to perceive the augmented visual effect, greatly reducing immersion. To track a moving viewer, an additional camera is placed facing in the opposite direction to the projector. However, view-agnostic methods typically use SfM to calibrate ProCams, recovering the scene only up to an unknown global scale. This scale ambiguity makes it difficult to register the oppositely facing tracking camera in the same world coordinate system. Furthermore, these methods rely on per-image iterative optimization, introducing substantial latency when compensation images must be continuously updated as the viewer moves. These limitations motivate the design of VAComp.

In this paper, we introduce \textbf{VAComp}, which dynamically updates the projector input image according to the viewer's viewpoint so that the observed projection matches the viewer virtual object image as the viewer moves. Using metric calibration as a reference, we recover the global scale of the SfM reconstruction and register the oppositely facing tracking camera using two checkerboards, one facing the RGB camera and the other facing the depth camera, as illustrated in \cref{fig:teaser}. This establishes a unified coordinate system for viewer tracking and view-dependent rendering. We further develop two non-iterative compensation methods, VAComp-SD and VAComp-SI, combining Gaussian-based ProCams rendering and geometric correction~\cite{deng2025} with learned photometric compensation networks~\cite{Wang2023CompenHREF,Li2026SetupIndependentFP}. Both methods generate each compensation image through a single forward pass. VAComp-SD supports real-time updates, while VAComp-SI eliminates per-setup photometric-network training and operates at interactive rates.

Our contributions can be summarized as follows:
 \begin{itemize}
  \item To the best of our knowledge, VAComp is the first framework to achieve real-time view-aware full projector compensation for a moving viewer in a naked-eye setting, without requiring glasses or other wearable displays, using only one projector.
  \item  We propose a method that calibrates a camera facing in the opposite direction from the projector and resolves the scale ambiguity in SfM reconstruction using the known metric distance between the projector and the RGB camera.
  \item We develop two non-iterative full projector compensation variants, VAComp-SD and VAComp-SI, that achieve competitive quality across novel viewpoints. VAComp-SD supports real-time updates, while VAComp-SI eliminates per-setup photometric network training and maintains interactive-rate performance.
\end{itemize}


\section{Related Work}
\label{sec:related_work}

\subsection{View-dependent projection mapping}

PM aims to seamlessly integrate virtual content with the physical world by projecting computer-generated imagery onto real-world surfaces. In recent years, PM has shown a clear trend toward increasingly dynamic settings. Dynamic Projection Mapping (DPM) has achieved low-latency, high-throughput projection across diverse targets, ranging from rigid objects~\cite{Halvorson2022Robust,Nomoto2020Realistic,Siegl2015Real,Watanabe2017ExtendedDC} to deformable surfaces~\cite{Narita2017Dynamic,Narita2015Dynamic,Ibrahim2022Dynamic} or deformable human body surfaces~\cite{Miyashita2018MIDAS,Erel2024Casper,Peng2025Perceptually,Zhang2026ImpressionGuidedIP,Zhang2026Text2Makeup}, while view-dependent projection mapping instead adapts the imagery to changes in the viewer's viewpoint. Prior work has demonstrated impressive visual realism in SAR and generally follows two directions: modifying the appearance of real objects by altering attributes such as color or texture~\cite{raskar2001shader,Lee2015Real,Siegl2015Real,Fujimoto2014Geometrically,Nomoto2020Realistic,Erel2024Casper,Zhang2026Text2Makeup,Raskar1999Table-top}, or augmenting the scene with virtual objects that are not physically present in the underlying geometry~\cite{Raskar1998office,Raskar1999Spatially,KhuslenBattulga2021,Fender2018OptiSpace,Bimber2005Enabling}.

The foundations of view-dependent projection mapping can be traced back to Raskar \etal's Office of the Future~\cite{Raskar1998office}, which envisioned turning everyday irregular surfaces into spatially immersive displays. To maintain perspective-correct imagery for a moving viewer, they introduced a two-pass projective-texture rendering scheme~\cite{Raskar1998EfficientIG}. Over the following two decades, this paradigm was extensively explored and extended in projector-camera systems, with representative developments comprehensively surveyed by Bimber \etal~\cite{Bimber2008} and Grundhöfer and Iwai~\cite{Grundhöfer2018}, and more recent advances reviewed by Iwai~\cite{Iwai2024}. Among representative systems from this period, Bimber \etal~\cite{Bimber2005Enabling} enabled view-dependent stereoscopic projection using an image-based warping scheme together with per-pixel radiometric compensation based on pre-calibrated camera-projector measurements. However, their stereoscopic visualization requires active shutter glasses. Siegl \etal~\cite{Siegl2015Real} combined multiple projectors and a depth camera with real-time pixel-luminance optimization, maintaining realistic PM even as the target object moved, while also supporting view-dependent projection mapping for a moving viewer. Nomoto \etal~\cite{Nomoto2020Realistic} introduced path tracing into DPM, producing highly realistic material and optical effects while adapting to both object motion and the viewer's viewpoint. Battulga and Fujimoto~\cite{KhuslenBattulga2021} addressed the empty regions exposed by shape differences between real and virtual objects by filling them with view-dependent background reprojection, making the virtual object appear more naturally integrated into the real scene.

These works demonstrate the strong potential of view-dependent projection mapping for compelling visual effects. However, many systems are demonstrated on white diffuse surfaces~\cite{Siegl2015Real,KhuslenBattulga2021}, while more advanced view-dependent applications, including stereoscopic visualization~\cite{Bimber2005Enabling,Okutani2018Stereoscopic}, material appearance rendering~\cite{Hamasaki2018HySAR}, and multifocal projection~\cite{Kimura2021Multifocal}, often require viewers to wear specialized hardware.

\begin{table}[tb]
    \caption{\textbf{Comparison of representative view-dependent projection mapping and projector compensation approaches with VAComp.} In the table, \fillcolor{red}{\ding{55}} and \fillcolor{green}{\ding{51}} indicate the limitations and capabilities of different methods, respectively. The first two methods are view-dependent
projection mapping approaches that adapt the projected content to the
viewer's viewpoint. Their photometric handling is nevertheless limited. Therefore, we use \fillcolor{yellow}{\ding{51}} to distinguish these methods from fully view-agnostic methods.}
     \label{tab:comparison_related_work}
    \centering  
    \setlength{\tabcolsep}{2.5pt}
    \resizebox{\linewidth}{!}{  
        \renewcommand{\arraystretch}{1.35}
        \begin{tabular}{|l|c|c|c|c|}
            \hline
            \textbf{Method}
            & \begin{tabular}[c]{@{}c@{}}Viewer-tracking\\method\end{tabular}
            & \begin{tabular}[c]{@{}c@{}}View-agnostic\\method\end{tabular}
            & \begin{tabular}[c]{@{}c@{}}Textured-surface\\compensation\end{tabular}
            & \begin{tabular}[c]{@{}c@{}}Real-time\\application\end{tabular} \\
            \hline
            Direct-light optimization~\cite{Siegl2015Real} &  \bggreen{\ding{51}} & \bgyellow{\ding{51}} & \bgred{\ding{55}} & \bggreen{\ding{51}} \\
            Homography-based~\cite{KhuslenBattulga2021} &  \bggreen{\ding{51}}   & \bgyellow{\ding{51}} & \bgred{\ding{55}} & \bggreen{\ding{51}}  \\  
            Setup-dependent CNN~\cite{Huang2021Full,Wang2023CompenHREF} & \bgred{\ding{55}} & \bgred{\ding{55}} & \bggreen{\ding{51}} & \bggreen{\ding{51}} \\
            Setup-independent CNN~\cite{Li2026SetupIndependentFP} & \bgred{\ding{55}}  & \bgred{\ding{55}} & \bggreen{\ding{51}} & \bgred{\ding{55}} \\
            Geometry-attribute shading~\cite{Huang2021DeProCams} & \bgred{\ding{55}} & \bgred{\ding{55}} & \bggreen{\ding{51}} & \bgred{\ding{55}}  \\
            Path-tracing~\cite{Li2025DPCS} & \bgred{\ding{55}} & \bgred{\ding{55}} & \bggreen{\ding{51}} & \bgred{\ding{55}}  \\
            3D representation-based~\cite{Erel2023Neural,deng2025} & \bgred{\ding{55}} & \bggreen{\ding{51}} & \bggreen{\ding{51}} & \bgred{\ding{55}}  \\
            VAComp (Ours) & \bggreen{\ding{51}} & \bggreen{\ding{51}} & \bggreen{\ding{51}} & \bggreen{\ding{51}} \\
            \hline
        \end{tabular}
    } 
\end{table}

\begin{figure*}[t]
    \centering
    \includegraphics[width=\textwidth]{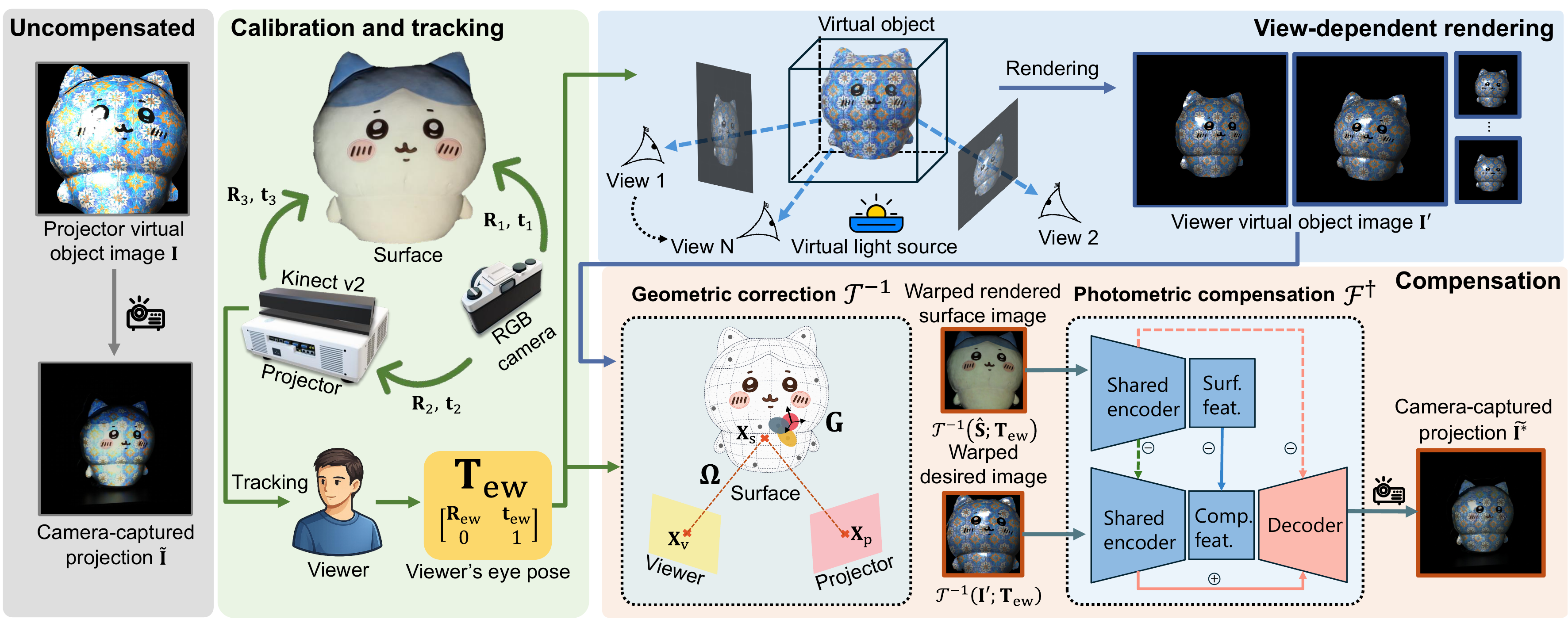}
    \caption{\textbf{View-aware full projector compensation (VAComp) pipeline.}
The uncompensated image is obtained by rendering the virtual object from the projector viewpoint and directly projecting the rendered image onto the projection surface. For our method, we first calibrate the projector, Kinect v2, RGB camera, and projection surface (see \cref{sec:camera_calibration}), and train GS-ProCams~\cite{deng2025} as the scene-specific forward renderer and geometric compensator. At runtime, VAComp estimates the viewer's eye pose and renders the viewer virtual object image. We use the geometric correction of GS-ProCams~\cite{deng2025} to warp the desired image into the projector domain. According to the photometric compensation network used, GS-ProCams renders the corresponding surface-prior images, which are also geometrically corrected. The geometrically corrected desired image and surface priors are then processed by the corresponding network for photometric compensation. We use a generalized network diagram to represent both photometric compensation architectures: the pink dashed residual connection is present in PANet, whereas the green dashed residual connection is present in IVPCNet. In the diagram, ``Surf.'', ``Comp.'', and ``Feat.'' denote surface, compensated, and feature, respectively. VAComp-SD employs PANet from CompenHR~\cite{Wang2023CompenHREF}, whereas VAComp-SI employs IVPCNet from SIComp~\cite{Li2026SetupIndependentFP}.}
    \label{fig:pipeline}

\end{figure*}

\subsection{Projector compensation}
Projector compensation aims to pre-correct the projector input image to counteract the undesirable visual effects when projecting onto non-planar and textured surfaces. The process of projector compensation typically comprises geometric correction~\cite{RaskariLamps2003,BoroomandSLCW16,Narita2015Dynamic,Raskar2001Self-correcting,Tardif2003Multi} and radiometric compensation~\cite{Wetzstein2007Raidometric,Grossberg2004make,Juang2007PhotometricSO,Grundhofer2015Robust,Bokaris2014Photometric,Kurth2020Proceedings,ChenColor2008,Huang2019End-To-End,Huang2017Radiometric,Aliaga2012Fast,Ashdown2006Robust,Yoshida2003virtual}.

While earlier approaches used specially designed patterns for simultaneous geometric and radiometric adaptation~\cite{Park2008Simultaneous} or calibration~\cite{Shahpaski2017Simultaneous}, recent work shows a clear trend towards designing a single, unified deep neural network to model both transformation functions. Huang \etal~\cite{Huang2021Full} developed an end-to-end framework that integrates two key components: a coarse-to-fine process to compute the geometric warping field, and a network for photometric compensation. Wang \etal~\cite{Wang2023CompenHREF} effectively improved efficiency in high-resolution scenarios by introducing the pixel shuffle operation. Li \etal~\cite{Li2026SetupIndependentFP} introduced SIComp, the first setup-independent framework for full projector compensation, which decouples geometric and photometric correction through an online optical-flow module and a surface-prior-guided photometric network, enabling direct generalization to unseen ProCams setups without fine-tuning. Recent fully differentiable ProCams frameworks~\cite{Huang2021DeProCams,Erel2023Neural,Li2025DPCS,deng2025} formulate full projector compensation as an optimization problem that enforces consistency between the relighting output and the target appearance. Erel \etal~\cite{Erel2023Neural} proposed a view-agnostic ProCams simulation based on a neural reflectance field and enabled projector compensation for novel viewpoints. Deng \etal~\cite{deng2025} proposed a Gaussian Splatting (GS)-based, view-agnostic ProCams framework, enabling efficient and high-quality ProCams simulation and full projector compensation for novel viewpoints. However, their compensation stage still relies on iterative optimization and therefore cannot provide real-time compensation. In \cref{tab:comparison_related_work}, we analyze representative categories of projector compensation and view-dependent projection mapping approaches and discuss their key features to facilitate a clear comparison with our VAComp.


\section{Method}
\label{sec:method}
\begin{figure*}[t]
    \centering
    \includegraphics[width=0.9\textwidth]{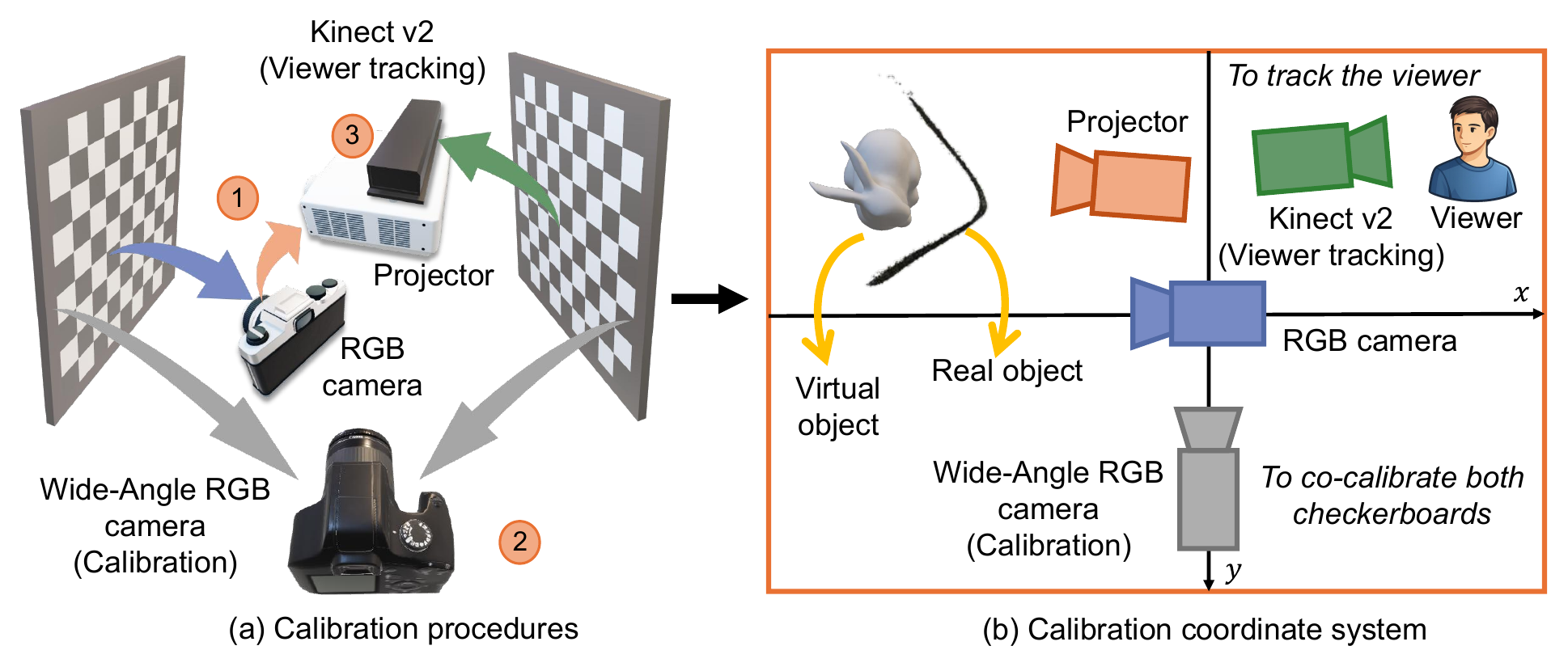}
    
    \caption{\textbf{Calibration pipeline.} (a) We first calibrate the RGB camera and the projector. We then place one checkerboard on the RGB-camera side and another on the side opposite to the projector’s projection direction. The wide-angle RGB camera is positioned such that both checkerboards are visible, while the Kinect v2 is placed on the side opposite the projector with a clear view of the corresponding checkerboard. We then capture images using the RGB camera, the Kinect v2, and the wide-angle RGB camera. The numbers in the figure indicate the order of the calibration steps. (b) Following (a), we register the current RGB-camera pose to the COLMAP reconstruction and recover its metric scale. The resulting world coordinate system is defined with the RGB-camera pose in (a) as the origin, in which the virtual object is placed. }
    \label{fig:calibration_pipeline}

\end{figure*}

\subsection{Problem formulation}
The primary goal of VAComp is to remove the static-viewpoint assumption by tracking the viewer in real time and altering the projector input such that, when projected onto colored and textured surfaces, the perceived projection closely matches the viewer virtual object image.

To describe this challenge, we model the physical process of projecting a virtual object into a real-world scene. The most straightforward approach is to render the virtual object from the projector viewpoint to obtain an image $\mathbf{I}$, and directly project it onto the projection surface without compensation, which we refer to as the uncompensated image $\tilde{\mathbf{I}}$. However, the appearance perceived by the viewer can deviate substantially from the intended virtual object. This distortion mainly arises from three factors: \textbf{viewpoint discrepancy}, since the virtual object is not rendered according to the viewer's current eye pose $\mathbf{T}_{\mathrm{ew}}$ where the subscripts \(\mathrm{e}\) and \(\mathrm{w}\) denote the viewer eye and world coordinate systems, respectively, and \(\mathbf{T}_{\mathrm{ew}}\) maps coordinates from the former to the latter; \textbf{geometric transformation} $\mathcal{T}$, which accounts for the
geometric warping induced by the surface geometry; and \textbf{photometric transformation} $\mathcal{F}$, which models the interaction between the projected light, the surface properties
$\boldsymbol{s}$ (geometry and reflectance), and the ambient
illumination $\boldsymbol{g}$. Thus, the uncompensated image $\tilde{\mathbf{I}}$ can be formulated as:
\begin{equation}
\tilde{\mathbf I}
=
\mathcal T
\left(
\mathcal F
\left(
\mathbf I;
\boldsymbol g,
\boldsymbol s,
\mathbf T_{\mathrm{ew}}
\right);
\mathbf T_{\mathrm{ew}}
\right).
\end{equation}

The purpose of VAComp is to find the compensation image $\mathbf{I}^{*}$ so that the final image captured by the viewer $\tilde{\mathbf{I}}^{*}$ matches the viewer virtual object image $\mathbf{I}'$:

\begin{equation}
\label{eq:cmp_forward}
  \mathcal T
\left(
\mathcal F
\left(
\mathbf{I}^{*};
\boldsymbol g,
\boldsymbol s,
\mathbf T_{\mathrm{ew}}
\right);
\mathbf T_{\mathrm{ew}}
\right)
=\tilde{\mathbf{I}}^{*} \approx \mathbf{I}'.
\end{equation}

However, directly measuring the complex effects of surface properties $\boldsymbol{s}$ and ambient illumination $\boldsymbol{g}$ is very difficult.
A common approach is to project a uniform gray image $\mathbf{I}_0$ onto
the surface and capture the resulting surface image $\tilde{\mathbf{S}}$,
which implicitly encodes their combined effects~\cite{Huang2021Full}. In our view-aware setting,
however, directly acquiring $\tilde{\mathbf{S}}$ from the viewer's current
viewpoint is infeasible because it would require repeated capture as the viewer
moves. We therefore employ the trained GS-ProCams~\cite{deng2025} forward renderer
$\mathcal{R}_{\theta}$, parameterized by the learned parameters $\theta$, to synthesize the corresponding viewer-captured surface
image from the current eye pose $\mathbf{T}_{\mathrm{ew}}$. The rendered surface
image $\hat{\mathbf{S}}$ serves as an approximation to the viewer-captured
surface image:
\begin{equation}
\mathcal{R}_{\theta}
\left(
\mathbf{I}_0,
\mathbf{T}_{\mathrm{ew}}
\right)
=
\hat{\mathbf{S}}
\approx
\tilde{\mathbf{S}}.
\end{equation}

To derive the compensation image $\mathbf{I}^{*}$, we aim to approximate
the inverse of the composite transformation in \cref{eq:cmp_forward}. However, obtaining
an exact inverse of the photometric transformation $\mathcal{F}$ is generally nontrivial due to the complex nonlinear light interactions between the
projector, projection surface, and environment. We therefore use $\mathcal{F}^{\dagger}$ as an approximation of the inverse photometric mapping. Using the rendered surface image
$\hat{\mathbf{S}}$ as the view-dependent surface prior, the compensation
image can be expressed as:
\begin{equation}
\mathbf{I}^{*}
=
\mathcal{F}^{\dagger}
\left(
\mathcal{T}^{-1}
\left(
\mathbf{I}'; \mathbf{T}_{\mathrm{ew}}
\right),
\mathcal{T}^{-1}
\left(
\hat{\mathbf{S}}; \mathbf{T}_{\mathrm{ew}}
\right)
\right),
\end{equation}
where $\mathcal{F}^{\dagger}$ serves as an approximation to the inverse photometric transformation, and $\mathcal{T}^{-1}$ maps images from the current viewer coordinate space to the projector coordinate space. The overall pipeline is illustrated in \cref{fig:pipeline}.

\begin{figure*}[t] 
\centering 
\includegraphics[width=\textwidth]{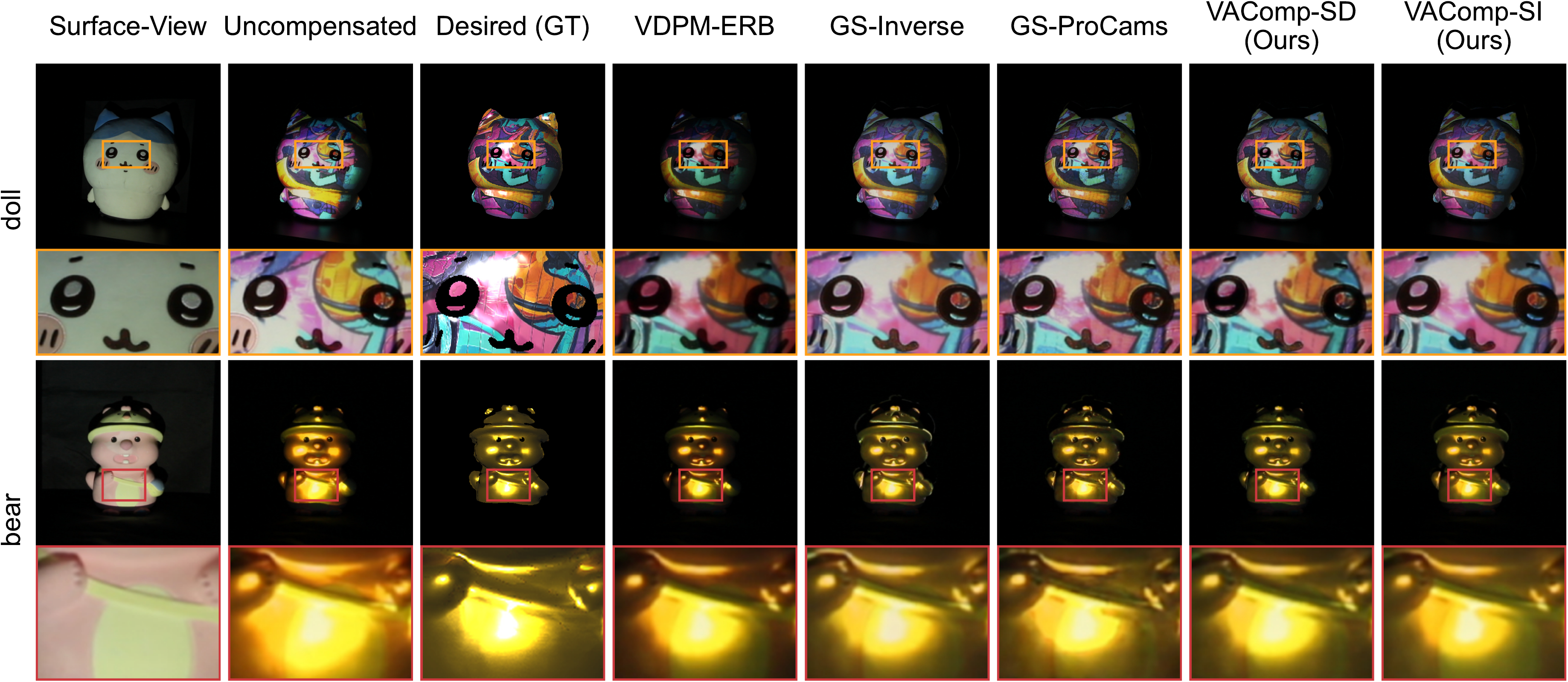}
\caption{\textbf{Visual comparisons of projector compensation for novel projections.} Each row represents a different surface, with the label at the beginning of the row indicating the name of the corresponding case. The first two columns show the camera-captured projection surfaces and their uncompensated images, while the remaining columns present the viewer virtual object images together with the camera-captured compensated projections produced by different methods. For better visualization, the images in the second row have been cropped and enlarged.}
\label{fig:comparison_methods}
\end{figure*}

\begin{table*}[t]
\centering
\caption{\textbf{Quantitative comparison of projector compensation on novel viewpoints with novel projections.}
Results are averaged over eight experimental cases derived from five different scenes. Each case contains 78 novel viewpoints. Client frames per second (FPS) is the client-side display refresh rate, and Repeat Ratio is the average number of repeated display frames per fresh compensation image. GS Train Time and PCN Train Time report the training times of the GS model and photometric compensation network (PCN) in minutes, respectively. In the PCN Train Time column, ``-'' indicates that the method does not use a PCN, whereas zero indicates that no per-setup PCN training is required. The best and second-best results are highlighted in \textbf{bold} and \underline{underlined}, respectively.}
\label{tab:comparison_methods}
\setlength{\tabcolsep}{3.8pt}
\begin{tabular*}{\textwidth}{@{\extracolsep{\fill}}lrrrrrrrr@{}}
\toprule
\textbf{Method}
& \textbf{PSNR}$\uparrow$
& \textbf{SSIM}$\uparrow$
& \textbf{LPIPS}$\downarrow$
& \textbf{$\Delta E$}$\downarrow$
& \textbf{Client FPS}$\uparrow$
& \textbf{Repeat Ratio}$\downarrow$
& \textbf{GS Train Time}$\downarrow$
& \textbf{PCN Train Time}$\downarrow$ \\
\midrule
VDPM-ERB~\cite{KhuslenBattulga2021}
& 14.3318
& 0.4340
& 0.3572
& 16.9982
& 42.35
& \textbf{0 repeats/frame}
& \underline{11.76~min}
& - \\
GS-Inverse
& 20.2499
& 0.6127
& 0.3190
& \underline{9.4547}
& \underline{42.87}
& \textbf{0 repeats/frame}
& \textbf{11.54~min}
& - \\
GS-ProCams~\cite{deng2025}
& 19.2644
& 0.6003
& 0.3138
& 9.6706
& 41.13
& 85 repeats/frame
& 11.91~min
& - \\
VAComp-SD (Ours)
& \underline{20.4819}
& \textbf{0.6403}
& \textbf{0.3032}
& \textbf{9.1447}
& \textbf{42.95}
& \textbf{0 repeats/frame}
& \underline{11.76~min}
& \underline{8.13~min} \\
VAComp-SI (Ours)
& \textbf{20.5432}
& \underline{0.6270}
& \underline{0.3122}
& 9.8710
& 42.48
& \underline{4 repeats/frame}
& \underline{11.76~min}
& \textbf{0.00~min} \\
\midrule
Uncompensated
& 15.1919
& 0.4603
& 0.3828
& 16.4294
& -
& -
& -
& - \\
\bottomrule
\end{tabular*}
\end{table*}

\subsection{Camera calibration}
\label{sec:camera_calibration}
In this section, we describe the camera calibration procedure adopted in our system. The calibration process consists of two steps: estimating the extrinsic parameters of our depth camera and resolving the scale ambiguity inherent in the SfM-based reconstruction (implemented using COLMAP~\cite{Schönberger2016COLMAP}). The overall calibration pipeline is illustrated in \cref{fig:calibration_pipeline}.

\noindent
\textbf{Extrinsic calibration of the depth camera.} In our system, the Kinect v2 tracks the viewer’s eye pose. Due to the opposite viewing directions of the Kinect v2 and the projector, the Kinect v2 and the RGB camera cannot simultaneously observe a single calibration checkerboard. To address this issue, we place two calibration checkerboards in the scene, each oriented toward one of the cameras.

As shown in \cref{fig:calibration_pipeline}, we first calibrate the RGB camera and the projector to estimate their extrinsic parameters in metric scale. We then place one checkerboard on the RGB-camera side and another checkerboard on the side opposite to the projector’s projection direction. Next, we position the wide-angle RGB camera so that it can observe both checkerboards and place the Kinect v2 on the side opposite the projector so that it can observe the corresponding checkerboard. We subsequently capture checkerboard images using the RGB camera, the Kinect v2, and the wide-angle RGB camera. Using these images, we first define the initial pose of the RGB camera as the reference frame of the world coordinate system. We then perform checkerboard-based calibration~\cite{Zhang2020CameraCalibration} to estimate the poses of the wide-angle RGB camera and the Kinect v2 relative to the initial RGB camera pose. Overall, this process yields the extrinsic parameters of the Kinect v2 facing in the opposite direction from the projector, thereby determining its pose in the world coordinate system.

\noindent
\textbf{Scale ambiguity resolution.} 
Since the Kinect extrinsics must be calibrated in metric space, the resulting calibration is defined in a real-world coordinate system with a known scale (in millimeters). However, SfM reconstructions are only determined up to an unknown global scale. This scale ambiguity introduces significant challenges when establishing a unified world coordinate system across cameras. Because SfM preserves relative geometry despite this scale ambiguity, our goal is to recover a global scale factor that aligns the SfM reconstruction with the metric coordinate system. We first calibrate the RGB camera-projector extrinsics in metric space and, following~\cite{deng2025}, perform SfM using multiple views of projected speckle patterns. We then scale the reconstruction by the ratio between the metric RGB camera-projector distance and its corresponding distance in the SfM reconstruction.

\subsection{Viewer's eye pose estimation}
\label{sec:viewer_eye_pose}
We model the viewer’s eye as a pinhole camera and approximate its intrinsic parameters using those of the calibrated RGB camera. In this section, we therefore focus on estimating the viewer’s eye pose $\mathbf{T}_{\mathrm{ew}}$ in the world coordinate system.

Let $P$ be a point in the world coordinate system. We denote its coordinates in the Kinect v2 coordinate system as $\mathbf{p}_{\mathrm{k}}$ and its coordinates in the viewer eye coordinate system as $\mathbf{p}_{\mathrm{e}}$. The relationship between these two representations is given by
\begin{equation}
\label{eq:eye_to_kinect}
\mathbf{p}_{\mathrm{k}} = \mathbf{R}_{\mathrm{ek}}\mathbf{p}_{\mathrm{e}} + \mathbf{t}_{\mathrm{ek}},
\end{equation}
where $\mathbf{R}_{\mathrm{ek}}$ and $\mathbf{t}_{\mathrm{ek}}$ denote the rotation and translation from the viewer coordinate system to the Kinect v2 coordinate system. Assuming the viewer looks at the object’s geometric center, the viewing direction is defined by the vector from the eye position to the object center. Thus, once the viewer’s 3D position in the Kinect coordinate system is known, the viewing orientation is determined using a fixed reference up vector.

As described in \cref{sec:camera_calibration}, the extrinsic parameters of the Kinect v2 with respect to the world coordinate system have been determined. Accordingly, a 3D point $P$ with coordinates $\mathbf{p}_{\mathrm{k}}$ in the Kinect v2 coordinate system can be transformed into the world coordinate system $\mathbf{p}_{\mathrm{w}}$ via
\begin{equation}
\label{eq:kinect_to_world}
\mathbf{p}_{\mathrm{w}} = \mathbf{R}_{\mathrm{kw}}\mathbf{p}_{\mathrm{k}} + \mathbf{t}_{\mathrm{kw}},
\end{equation}
where $\mathbf{R}_{\mathrm{kw}}$ and $\mathbf{t}_{\mathrm{kw}}$ denote the rotation matrix and translation vector that transform points from the Kinect v2 coordinate system to the world coordinate system, respectively. By substituting \cref{eq:eye_to_kinect} into \cref{eq:kinect_to_world}, we can express the coordinates of point $P$ in the world coordinate system directly in terms of its coordinates in the viewer eye coordinate system:
\begin{equation}
\mathbf{p}_{\mathrm{w}} = \mathbf{R}_{\mathrm{kw}}\mathbf{R}_{\mathrm{ek}}\mathbf{p}_{\mathrm{e}} + \mathbf{R}_{\mathrm{kw}}\mathbf{t}_{\mathrm{ek}} + \mathbf{t}_{\mathrm{kw}}.
\end{equation}
This expression corresponds to the eye-to-world transformation, which maps points from the viewer eye coordinate system to the world coordinate system. Accordingly, the rotation and translation of this transformation are given by
\begin{equation}
\mathbf{R}_{\mathrm{ew}} = \mathbf{R}_{\mathrm{kw}}\mathbf{R}_{\mathrm{ek}},
\end{equation}
\begin{equation}
\mathbf{t}_{\mathrm{ew}} = \mathbf{R}_{\mathrm{kw}}\mathbf{t}_{\mathrm{ek}} + \mathbf{t}_{\mathrm{kw}},
\end{equation}
where $\mathbf{R}_{\mathrm{ew}}$ and $\mathbf{t}_{\mathrm{ew}}$ denote the rotation matrix and translation vector from the viewer eye coordinate system to the world coordinate system, respectively. The viewer's eye pose $\mathbf{T}_{\mathrm{ew}}$ can be compactly represented in homogeneous coordinates as
\begin{equation}
\label{eq:viewer_pose}
\mathbf{T}_{\mathrm{ew}} =
\begin{bmatrix}
  \mathbf{R}_{\mathrm{ew}}& \mathbf{t}_{\mathrm{ew}}\\
  0& 1
\end{bmatrix}.
\end{equation}

\subsection{VAComp: view-aware full projector compensation}
In this section, we present the modeling of view-aware full projector compensation. Specifically, this section consists of two parts. We first explain how the ideal view of the virtual object is rendered according to the viewer’s eye pose. We then describe how the compensation image is generated in our system so that the final projected appearance matches the viewer virtual object image.

As shown in \cref{fig:calibration_pipeline}, we first place the 3D virtual object in a virtual environment and adjust it to an appropriate scale. At runtime, we render the virtual object from the viewer's viewpoint to obtain the viewer virtual object image, which serves as the ground-truth target for view-aware full projector compensation.
\begin{table}[tb]
\centering
\caption{\textbf{Latency performance of the proposed framework.} For each case-method pair, the stage-wise processing times are averaged over 1,000 display refresh frames. The reported values are then averaged across all evaluated methods and eight cases.}
\label{tab:latency_performance}
{\setlength{\tabcolsep}{4pt}
\begin{tabular}{@{}lrr@{}}
\toprule
\textbf{Process} & \textbf{Processing time (ms)} & \textbf{Percentage} \\
\midrule
Eye tracking              & 8.5  & 36\% \\
Compensation-image update & 14.5 & 61\% \\
Rendering virtual object  & 0.8  & 3\%  \\
\midrule
Total latency             & 23.8 & -   \\
\bottomrule
\end{tabular}}
\end{table}

Existing view-agnostic full projector compensation methods based on ProCams
simulation~\cite{Erel2023Neural,deng2025} support compensation for novel
viewer poses through iterative optimization. Although effective, the repeated
optimization steps incur substantial computational overhead and introduce
prohibitive latency when compensation must be continuously updated as the
viewer moves. To improve runtime efficiency, we therefore seek a single-pass
view-aware full projector compensation framework that predicts the compensation image conditioned on the current viewer pose without per-frame iterative
optimization.

For geometric correction, inspired by GS-ProCams~\cite{deng2025},
we leverage its geometry-based warping component. As illustrated in \cref{fig:pipeline}, given the current viewer eye pose $\mathbf{T}_{\mathrm{ew}}$ and the reconstructed
projection-surface geometry $\mathbf{G}$, we first compute the corresponding
dense warping field as
\begin{equation}
\boldsymbol{\Omega}
=
\mathcal{D}
\left(
\mathbf{G}, \mathbf{T}_{\mathrm{ew}}
\right).
\end{equation}
Here, $\mathcal{D}$ projects the reconstructed projection-surface geometry
into the current viewer view and establishes a dense correspondence between
the viewer and projector domains.

For photometric compensation, as shown in \cref{fig:pipeline}, we employ a
photometric compensation network to approximate the inverse photometric
mapping $\mathcal{F}^{\dagger}$. Using the dense warping field
$\boldsymbol{\Omega}$, both the viewer virtual object image $\mathbf{I}'$
and the rendered surface image $\hat{\mathbf{S}}$ are geometrically
transformed into the projector domain. $\mathcal{W}(\cdot;\boldsymbol{\Omega})$ denotes the image-warping
operator parameterized by the dense warping field $\boldsymbol{\Omega}$,
which maps an image from the viewer domain to the projector domain.  The final compensation image is expressed as
\begin{equation}
\mathbf{I}^{*}
=
\mathcal{F}^{\dagger}
\left(
\mathcal{W}
\left(
\mathbf{I}'; \boldsymbol{\Omega}
\right),
\mathcal{W}
\left(
\hat{\mathbf{S}}; \boldsymbol{\Omega}
\right)
\right).
\end{equation}

Based on this shared framework, we instantiate two variants with different photometric compensation networks. VAComp-SD adopts PANet from CompenHR~\cite{Wang2023CompenHREF}, which requires per-setup photometric-network training data captured from a frontal reference view and geometrically rectified using structured light for network training. In contrast, VAComp-SI employs IVPCNet from SIComp~\cite{Li2026SetupIndependentFP}, whose setup-independent design enables direct deployment without per-setup photometric retraining. In our system, the computational cost of IVPCNet becomes a runtime bottleneck when VAComp-SI is integrated into the complete system. We adopt an asynchronous client-server architecture~\cite{Erel2024Casper} and offload compensation inference to a GPU server. The client performs viewer tracking and
view-dependent rendering, exchanges data with the server over Ethernet, and
displays the latest returned compensation image without blocking the local
processing loop.


\begin{figure}[tb]
    \centering
    \includegraphics[width=\columnwidth]{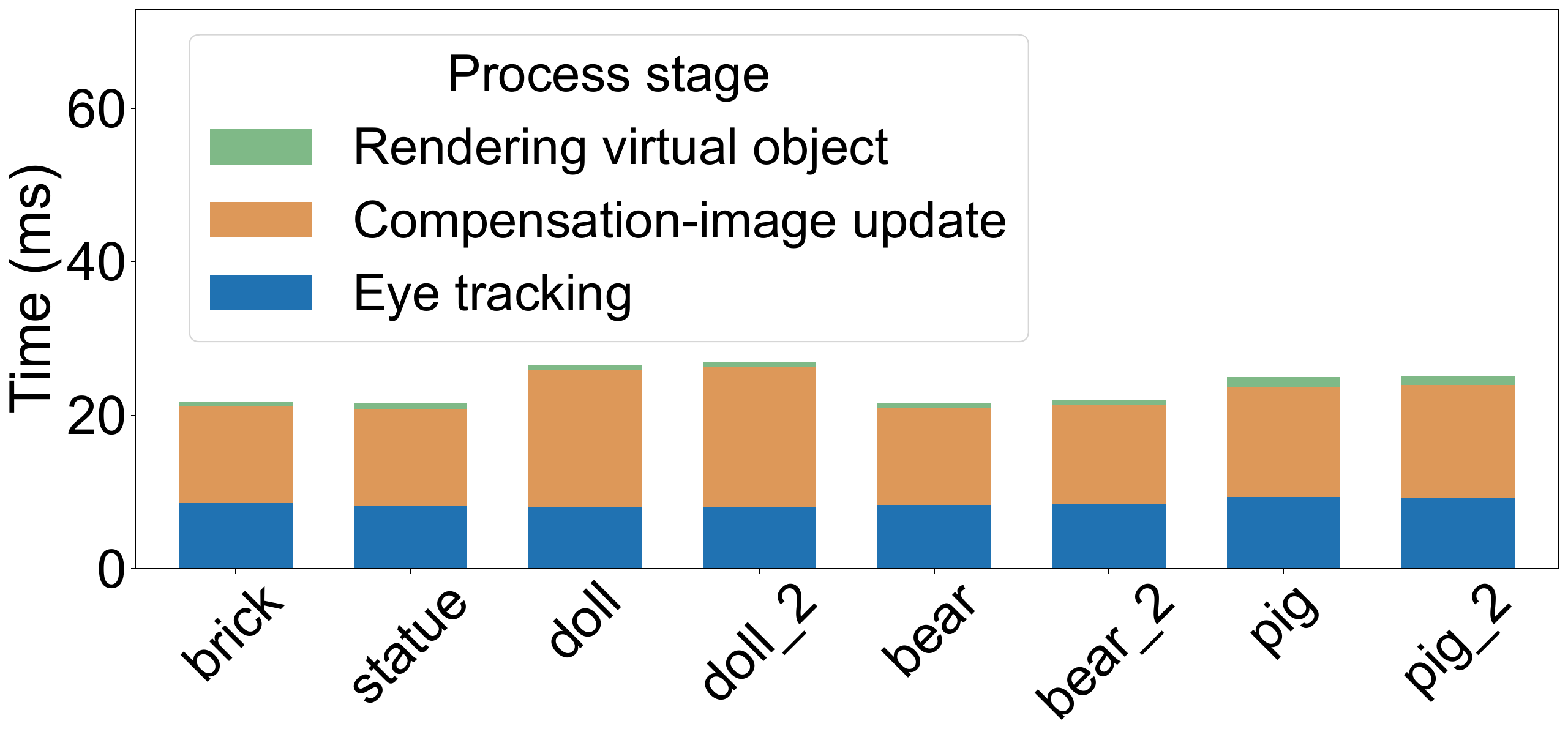}
    \caption{\textbf{Client-side latency breakdown.} The x-axis represents the eight evaluated cases, and the y-axis shows the processing time in milliseconds. The blue, orange, and green portions represent eye tracking, compensation-image update, and virtual object rendering, respectively. For cases whose qualitative results are not presented in the main paper, the corresponding results are provided in the supplementary material.}
    \label{fig:latency_breakdown}
\end{figure}
\begin{figure*}[t]
    \centering
    \includegraphics[width=\textwidth]{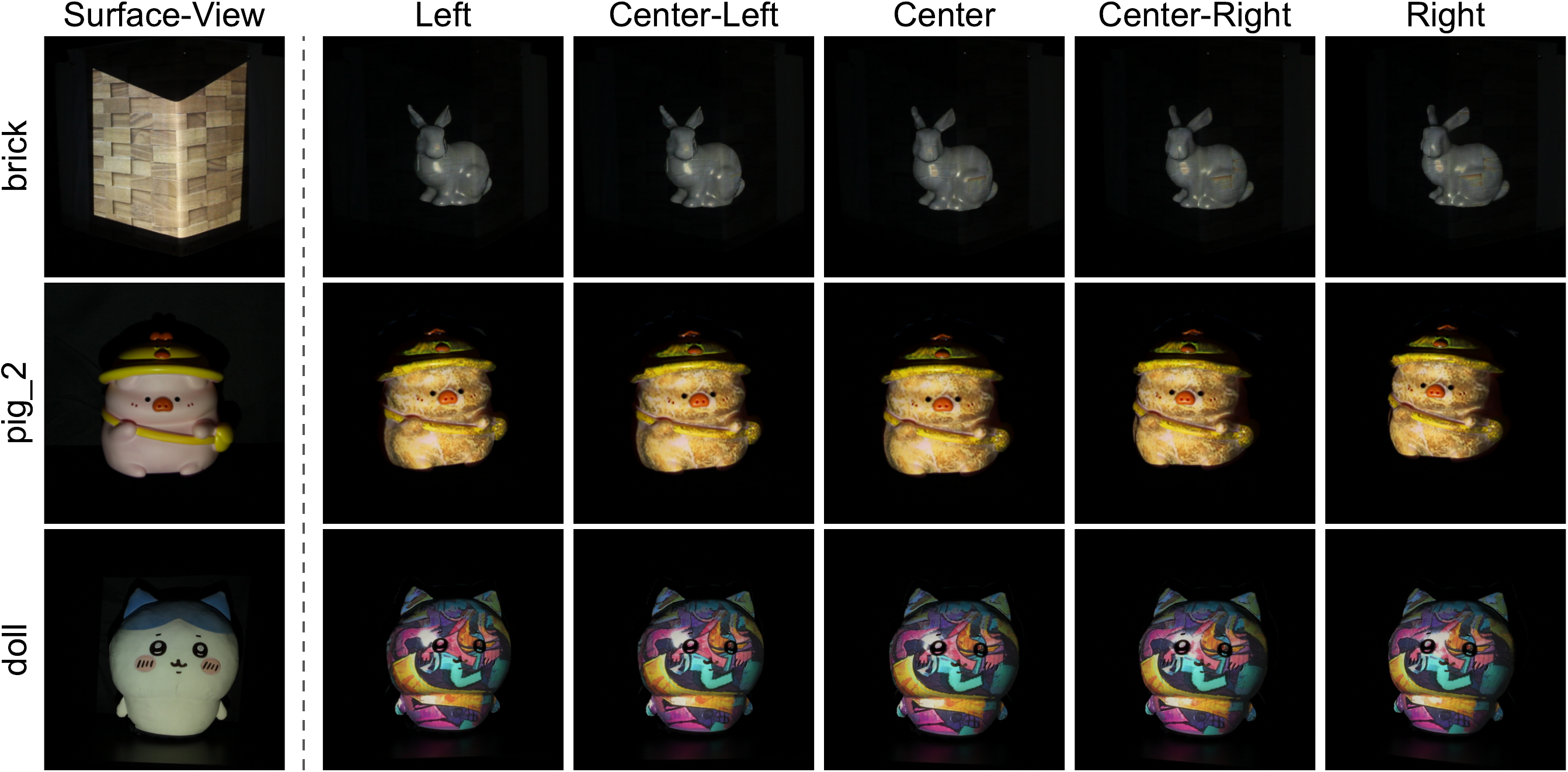}
   
    \caption{\textbf{Augmented views under multiple viewpoints.} Each row corresponds to the augmented result on a different projection surface, with the label on the left indicating the name of the corresponding case. From left to right, the first column shows the front-view camera-captured projection surfaces, while the remaining columns show the camera-captured projections under different viewpoints. The first row demonstrates the integration of a virtual object into a real scene with complex geometry and surface texture. The remaining rows illustrate how our framework enhances the appearance of existing real objects through material modification. All camera-captured compensated projections shown in these results are
produced using compensation images generated by VAComp-SD. For clearer visualization, the images in the first two rows are cropped and enlarged.
    }
     \label{fig:multiview_results}
\end{figure*}

\begin{figure}[tb]

    \centering
    \includegraphics[width=\columnwidth]{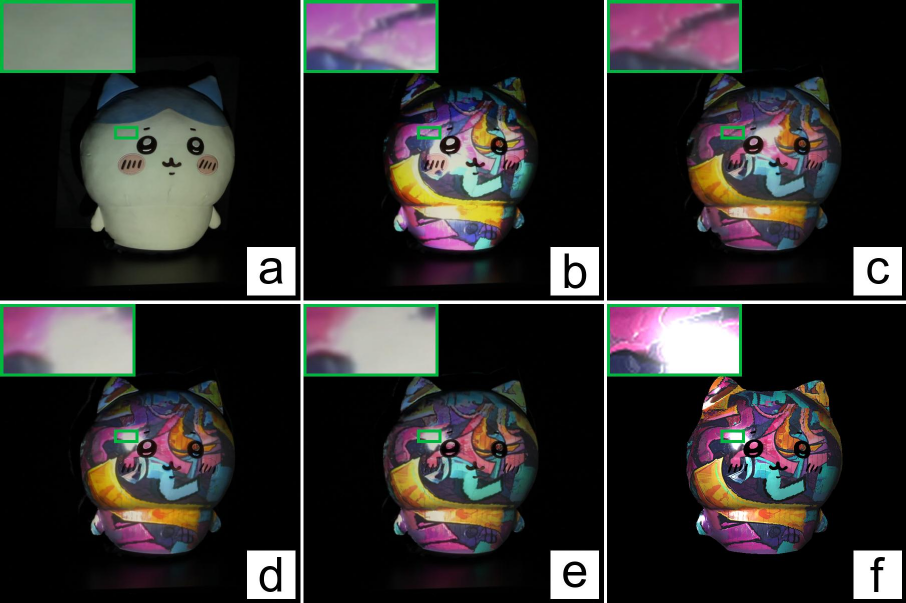}

    \caption{\textbf{Qualitative comparison of different projection strategies under viewpoint change for the first SAR application (real-object appearance enhancement).} The images are captured at a specific viewpoint during multi-view motion. (a) Camera-captured projection surface. (b) Camera-captured uncompensated projection. (c) Camera-captured projection using static compensation under viewpoint mismatch (right-side compensation, left-side capture). (d) Camera-captured compensated projection using VAComp-SI. (e) Camera-captured compensated projection using VAComp-SD. (f) Viewer virtual object image at this viewpoint.
    }
    \label{fig:comparison_appearance}
\end{figure}
\begin{figure}[tb]

    \centering
    \includegraphics[width=\columnwidth]{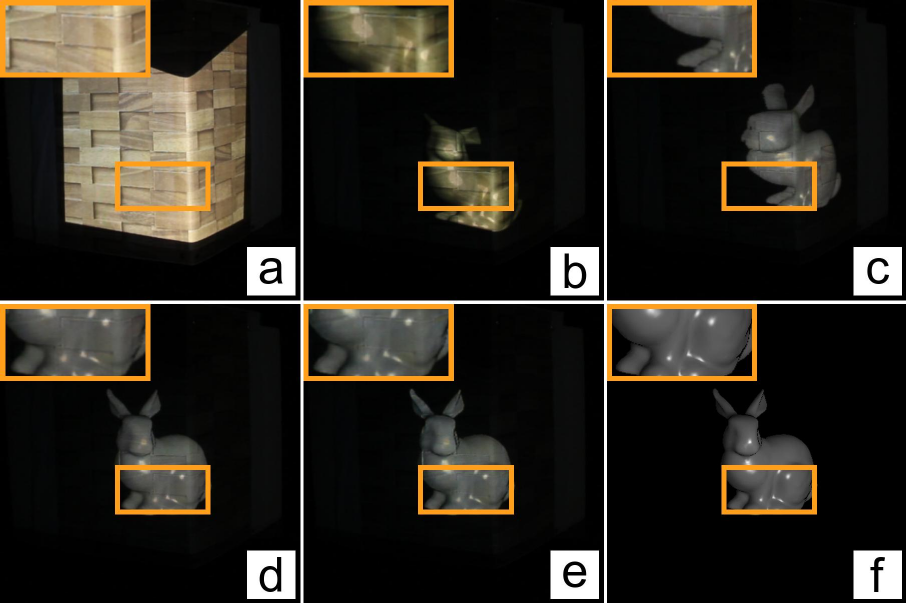}
   
    \caption{\textbf{Qualitative comparison of projection strategies under dynamic viewpoint change for the second SAR application (virtual-object replacement).} For clearer visualization, these images are cropped and enlarged. The images are captured at a specific viewpoint during multi-view motion. (a) Camera-captured projection surface. (b) Camera-captured uncompensated projection. (c) Camera-captured projection using static compensation under viewpoint mismatch (right-side compensation, left-side capture). (d) Camera-captured compensated projection using VAComp-SI. (e) Camera-captured compensated projection using VAComp-SD. (f) Viewer virtual object image at this viewpoint.
    }
    \label{fig:comparison_replacement}
\end{figure}
\begin{figure}[!htbp]
    \centering
    \includegraphics[width=\columnwidth]{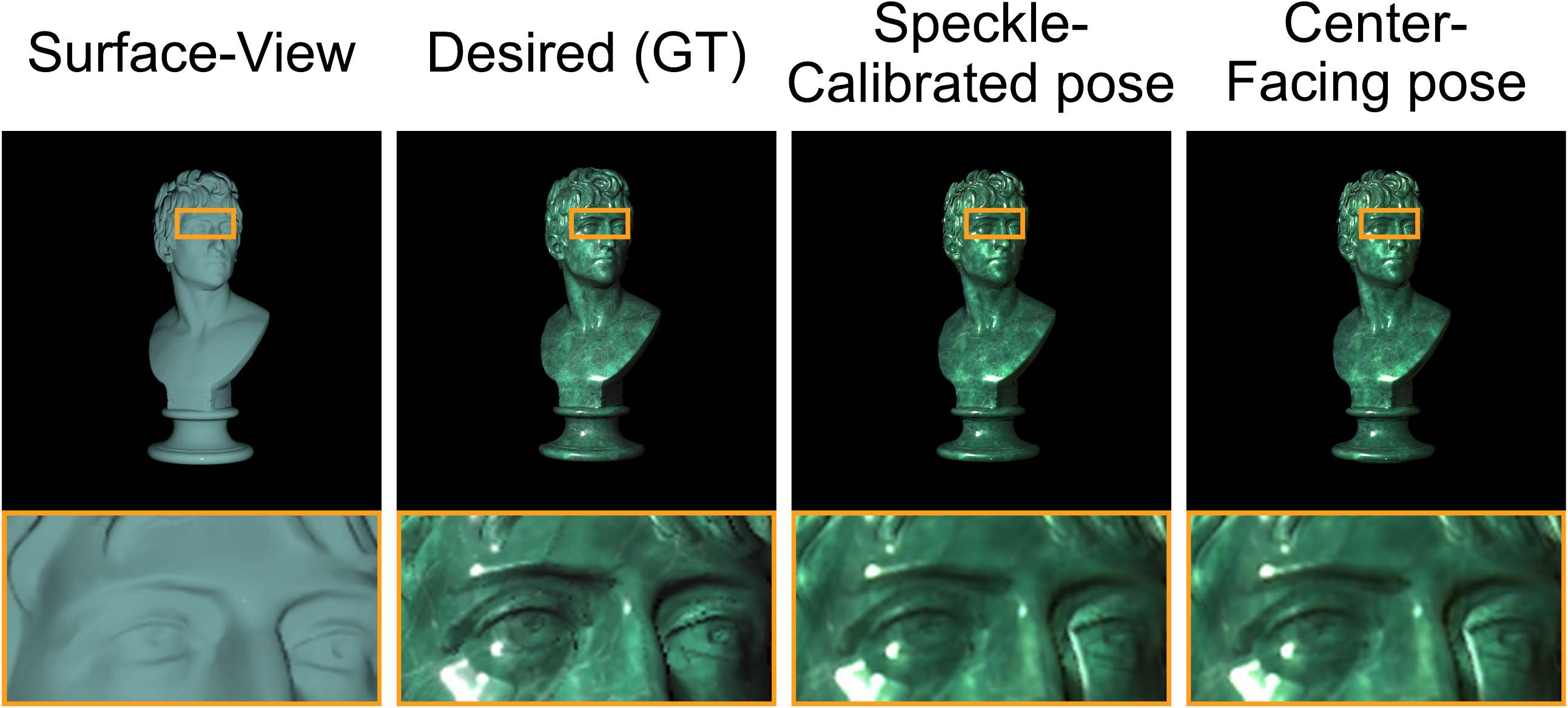}
    \caption{\textbf{Comparison of pose assumptions for projector compensation using VAComp-SI.} From left to right, the columns show the simulated projection surface, the viewer virtual object image, and the simulated compensated projections generated by VAComp-SI using the speckle-calibrated pose and the center-facing pose assumption, respectively. The bottom row shows magnified views of the regions highlighted by the orange boxes.}
    \label{fig:pose_comparison}
\end{figure}

\section{Experiments}
\label{sec:experiments}
\subsection{System configuration}

As shown in \cref{fig:teaser}, our ProCams setup consists of a projector, an RGB camera, and a Kinect v2. Our setup uses an Epson EB-C2025WN projector and a Canon EOS 600D DLSR camera. A Kinect for Windows v2 is also deployed facing the direction opposite to the projector to track the viewer's eye pose. In addition, an auxiliary wide-angle RGB camera is used exclusively during the calibration stage, as illustrated in \cref{fig:calibration_pipeline}.

We place our projector and cameras approximately 0.8-1.0~m away from the projection surface. The projector and RGB camera both operate at a resolution of 800~\texttimes~800, while the Kinect v2 captures images at 1920~\texttimes~1080. All models are trained on a laptop equipped with an Intel Core i7-14700HX processor, 16~GB RAM, and an NVIDIA GeForce RTX 4070 Laptop GPU with 8~GB GRAM. For dynamic compensation, our system adopts an asynchronous architecture that distributes different tasks across two computing devices. The full projector compensation algorithms are deployed on a server equipped with an NVIDIA GeForce RTX 3090 GPU with 24~GB GRAM, which is dedicated to generating compensation images. Meanwhile, the laptop performs eye tracking, virtual object rendering, and the display of the latest available compensation image.

\subsection{Comparison with state-of-the-art}

\label{sec:comparison_with_state_of_the_art}

To evaluate the effectiveness of our framework for real-time view-aware full projector compensation, we compare VAComp-SI and VAComp-SD with several representative baselines. We include GS-ProCams~\cite{deng2025}, a view-agnostic Gaussian-based ProCams simulation method. To obtain a stronger GS-ProCams baseline, we increase the maximum number of compensation iterations and introduce an early-stopping criterion based on peak signal-to-noise ratio (PSNR). We also compare with view-dependent projection mapping enhanced by real background (VDPM-ERB)~\cite{KhuslenBattulga2021}, which mitigates empty regions by projecting a view-dependent background. However, VDPM-ERB was not designed for full projector compensation and was not validated on textured surfaces with complex geometry. Following its view-dependent projection strategy, we therefore construct a stronger baseline using our geometric correction module while omitting photometric compensation and background synthesis, allowing us to evaluate the effect of geometric correction alone under the same experimental setting. To establish a view-agnostic real-time full projector compensation baseline, we construct GS-Inverse while keeping the same geometric correction module. Specifically, GS-Inverse performs inverse photometric compensation using the scene parameters learned by the GS-ProCams~\cite{deng2025} variant without Point Spread Function (PSF) modeling. Overall, the evaluated methods use a consistent geometric correction stage, while differing primarily in how photometric compensation is performed. All methods are evaluated under the same ProCams setup and lighting conditions.

To rigorously evaluate multi-view compensation performance, we conduct experiments across five distinct scenes comprising eight cases, with each case evaluated using camera-captured projections at 78 novel viewpoints distributed around the object. We adopt PSNR, Structural Similarity Index Measure (SSIM), Learned Perceptual Image Patch Similarity (LPIPS)~\cite{zhang2018unreasonable}, and perceptual color difference $\Delta E$~\cite{sharma2005ciede2000} for quantitative assessment. Beyond these quality metrics, we evaluate the computational cost of our approach during both training and dynamic compensation. For training, we report the training time as well as the average and peak GPU memory (GRAM) usage. For dynamic compensation, we measure the average and peak RAM and GRAM usage on both the laptop and the server. We also report the Client FPS, which measures the rate of the complete client-side loop, including eye tracking, virtual object rendering, compensation-result retrieval, and display. Since compensation is performed asynchronously, the client reuses the latest available result when no new compensation image has arrived. We additionally report the Repeat Ratio, defined as the average number of client display refreshes that reuse the previous compensation image for each newly received compensation image. A lower Repeat Ratio indicates more frequent compensation updates. In \cref{tab:comparison_methods}, we present only a subset of the computational performance results due to space limitations; the remaining results are provided in the supplementary material.

From the high Repeat Ratio reported in \cref{tab:comparison_methods}, we observe that GS-ProCams~\cite{deng2025} cannot provide fresh compensation images at the real-time display rate. This is mainly because generating each compensation image requires multiple iterative optimization steps. In contrast, both VAComp-SD and VAComp-SI predict each compensation image in a single forward pass without per-image iterative optimization. This efficiency does not come at the expense of compensation quality. By combining the high-fidelity rendering capability of the Gaussian scene representation with robust learned photometric compensation, both methods achieve a more favorable trade-off between update speed and compensation quality than GS-ProCams~\cite{deng2025}.

As shown in \cref{fig:comparison_methods}, VDPM-ERB exhibits substantial deviations from the viewer virtual object image on real objects with colored textures. Moreover, the quantitative results in \cref{tab:comparison_methods} show that applying geometric compensation alone can even yield lower compensation quality than Uncompensated. This behavior is expected because VDPM-ERB handles regions exposed by shape differences between the real and virtual objects, but does not perform photometric compensation. VDPM-ERB remains effective for objects with approximately white surfaces, such as white statues or foam objects, where surface-induced photometric effects are relatively limited. Our results instead highlight the importance of photometric compensation for view-aware settings: without it, the observed projection can deviate substantially from the viewer virtual object image, particularly on objects with colored textures.

In comparison, GS-Inverse performs full projector compensation and therefore achieves substantially better compensation quality than VDPM-ERB and Uncompensated, as shown in \cref{tab:comparison_methods} and \cref{fig:comparison_methods}. Nevertheless, it underperforms VAComp-SD and VAComp-SI because the reused GS-ProCams parameters are optimized for forward image formation rather than stable analytic inversion. By contrast, VAComp-SD and VAComp-SI employ learned photometric compensation networks. The results suggest that our models are more tolerant of modeling errors than the purely analytic inverse, contributing to their improved compensation quality.

VAComp-SI is slower than VAComp-SD because it employs a more complex photometric network architecture for setup independence. VAComp-SD achieves the best compensation quality, which is expected because its photometric network is optimized separately for each fixed setup. In contrast, VAComp-SI does not require setup-specific photometric network training, reducing per-setup data collection and training overhead. Without setup-specific photometric training, VAComp-SI still outperforms all other evaluated baselines in compensation quality. Overall, VAComp-SD provides the highest accuracy when setup-specific optimization is available, whereas VAComp-SI offers a favorable balance among compensation quality, generalization, and deployment efficiency.

\subsection{Latency performance}

We evaluate the client-side latency of our asynchronous system by decomposing each display refresh into three stages: eye tracking, virtual object rendering, and compensation-image update. Eye tracking comprises face detection, eye-pose estimation, and the coordinate transformations required to obtain the current viewpoint, while compensation-image update denotes retrieving and displaying the latest available compensation image.

For each case-method pair, we measure the processing time of each stage over 1,000 display refresh frames. The measurements are then averaged across all evaluated methods and eight cases. The overall average latency breakdown is reported in \cref{tab:latency_performance}, while the case-wise results, averaged across methods, are shown in \cref{fig:latency_breakdown}. Additional qualitative results for cases not visualized in the main paper are provided in the supplementary material.

As shown in \cref{tab:latency_performance}, compensation-image update accounts for the largest portion of the client-side processing time, requiring 14.5~ms per refresh, or 61\% of the total latency. Eye tracking takes 8.5~ms (36\%), whereas virtual object rendering requires only 0.8~ms (3\%). The resulting total latency is 23.8~ms per refresh, corresponding to a theoretical client refresh rate of approximately 42.0~FPS. Together with the Repeat Ratio reported in \cref{tab:comparison_methods}, these results show that VAComp-SD supports real-time compensation updates, while VAComp-SI maintains interactive update rates.

\subsection{Augmented views}
SAR generally serves two primary objectives. One is to \emph{enhance the appearance of a real object by projecting decorative or dynamic imagery onto its surface}~\cite{raskar2001shader,Lee2015Real,Siegl2015Real,Fujimoto2014Geometrically,Nomoto2020Realistic,Erel2024Casper,Zhang2026Text2Makeup,Raskar1999Table-top}. The other is to \emph{overlay a virtual object onto a real surface that does not inherently correspond to the virtual object}~\cite{Raskar1998office,Raskar1999Spatially,KhuslenBattulga2021,Fender2018OptiSpace,Bimber2005Enabling}. To demonstrate that our framework effectively supports both application scenarios, we conduct experiments across eight cases. For each case in \cref{sec:comparison_with_state_of_the_art}, we assemble all frames captured using each evaluated method into a video sequence. The supplementary video is available on the project page: \url{https://realkaizhou.github.io/VAComp/}. \cref{fig:multiview_results} presents representative frames arranged according to the viewpoint changes, while the supplementary video provides dynamic comparisons between VAComp and representative static compensation methods under changing viewpoints. \cref{fig:comparison_appearance,fig:comparison_replacement} present representative frames from these comparisons for the two application scenarios, intuitively highlighting the need for VAComp.

For the first application scenario (real-object appearance enhancement), the goal is to enhance the appearance of an existing real object. This requires a 3D model whose geometry is aligned with that of the real object~\cite{KhuslenBattulga2021}. If a geometrically matched 3D model of the target object is unavailable, we reconstruct one from the Gaussian representation. Following~\cite{Bingbing20242DGaussian}, we extract a geometrically aligned mesh from the Gaussian representation using truncated signed distance function (TSDF) fusion and refine it through smoothing and denoising, enabling material and appearance editing in the virtual environment. A virtual light source is placed in the scene to produce view-dependent illumination effects. In the second and third rows of \cref{fig:multiview_results}, we replace the original materials of the models with stylized materials. The virtual light source produces specular highlights on their surfaces, which shift accordingly as the camera moves from left to right. This view-dependent variation illustrates the appearance changes produced by the configured virtual materials and lighting as the viewpoint changes.

We first compare different projection strategies for the first application scenario (real-object appearance enhancement), as illustrated in the supplementary video and \cref{fig:comparison_appearance}. As shown in \cref{fig:comparison_appearance}, the camera-captured uncompensated projection deviates substantially from the viewer virtual object image. The static compensation method preserves the overall appearance of the stylized graffiti ceramic material at a new viewpoint, but fails to reproduce the view-dependent specular highlights present in the viewer virtual object image. As a result, the appearance remains visually static and does not respond naturally to viewpoint changes, diminishing the sense of realism. In contrast, our method reproduces viewpoint-dependent specular highlights consistent with the viewer virtual object image as the viewer moves.

For the second application scenario (virtual-object replacement), as shown in the first row of \cref{fig:multiview_results}, the shape of the projection surface differs from that of the virtual object. As the viewer moves from left to right, despite these differences in geometry and appearance, the projected result remains visually consistent with the viewer virtual object image at each viewpoint.

We further compare different projection strategies for the second application scenario (virtual-object replacement), as illustrated in the supplementary video and \cref{fig:comparison_replacement}. The captured uncompensated projection exhibits a substantial discrepancy from the viewer virtual object image. The static compensation method lacks awareness of the viewer's dynamic viewpoint. When the viewer moves to a new viewpoint, this method cannot generate the corresponding viewer virtual object image, and its geometric corrections and photometric compensation no longer match the new viewpoint. Consequently, noticeable geometric and photometric inconsistencies appear in the captured projection. In contrast, our framework updates both the viewer virtual object image and the compensation image according to the viewer's eye pose, preserving geometric and photometric consistency across viewpoints.


\section{Discussion}

This section first examines the validity of a core pose assumption in our framework and then discusses the remaining limitations of the current system.

\noindent\textbf{Validity of the center-facing pose assumption.}
Our framework approximates the viewer's gaze direction as pointing from the estimated eye position toward the center of the target object. Because the actual gaze direction may deviate from this approximation, we assess how this difference affects the resulting compensation. We conduct a controlled experiment in Blender using five virtual setups, with 10 test viewpoints per setup. We use a virtual environment because, in a real-world setup, it is difficult to precisely enforce and verify that the viewer's viewing direction points exactly toward the center of the target object, whereas Blender allows the camera position and orientation to be accurately controlled. At each viewpoint, we compare the simulated compensated projection generated using the center-facing pose with that generated using a reference pose obtained through the speckle-based SfM calibration procedure adopted from GS-ProCams~\cite{deng2025} and implemented using COLMAP~\cite{Schönberger2016COLMAP}. As illustrated in \cref{fig:pose_comparison}, the two sets of simulated compensated projections achieve an average PSNR of 42.4265~dB, SSIM of 0.9949, LPIPS of 0.0050, and $\Delta E$ of 0.2128 across the five experimental scenes. These results suggest that the center-facing assumption provides a practical pose approximation within the evaluated setups and viewpoint range. Its reliability under larger gaze deviations and real-world tracking errors, however, remains to be investigated. Beyond this pose assumption, the current system has three additional limitations.
\begin{figure}[tb]
    \centering
    \includegraphics[width=\columnwidth]{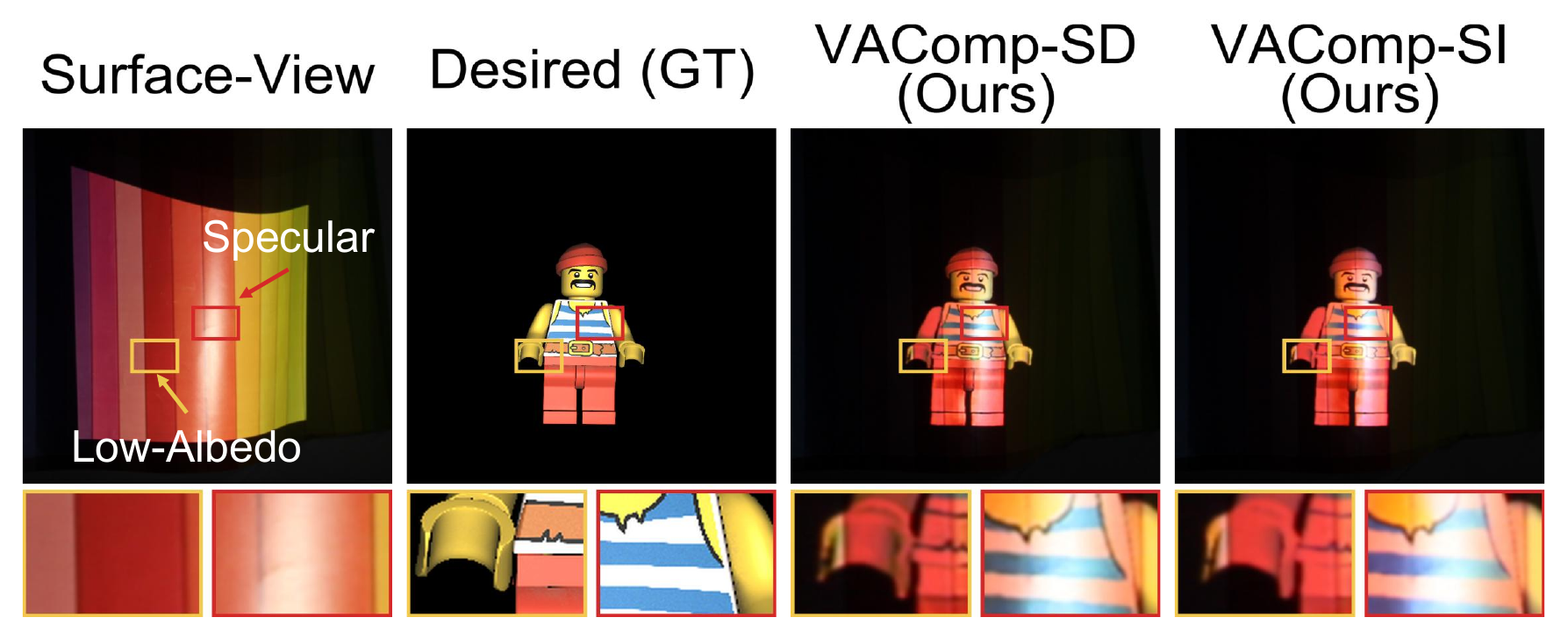}
    \caption{\textbf{Failure case.} VAComp-SD and VAComp-SI cannot reliably handle low-albedo regions or regions affected by specular highlights. These two cases respectively reflect the projector's limited ability to compensate for low surface reflectance and the difficulty of reproducing view-dependent reflections.}
    \label{fig:failure_cases}
\end{figure}

\noindent\textbf{Limitations on challenging surface appearances.}
As shown in \cref{fig:failure_cases}, both VAComp-SD and VAComp-SI struggle to compensate for low-albedo regions and view-dependent specular highlights, but for different reasons. In low-albedo regions, the limited output power and dynamic range of the projector can place the desired appearance outside its physically achievable gamut. For view-dependent specular highlights, GS-ProCams~\cite{deng2025}, which serves as our forward renderer, cannot provide surface-prior renderings that faithfully preserve such effects. Moreover, PANet is trained using observations from a single viewpoint and is therefore limited to colored diffuse surfaces without view-dependent specular highlights. We also observe degraded performance on plush toys with long, loosely distributed fibers, where complex light scattering among the fibers is not adequately captured by our current image-formation model.

\noindent\textbf{Reliance on setup-specific geometry.}
Although VAComp-SI does not require training a photometric compensation network, it relies on a GS representation optimized for each setup to provide scene geometry. Thus, the setup dependence of VAComp-SI is limited to scene geometry within the full projector compensation pipeline.

\noindent\textbf{Simplified modeling of the human visual system.}
In our framework, the viewer's eye is approximated by an RGB camera. However, the human eye differs from a camera in focal characteristics, dynamic range, color perception, and exposure adaptation. Consequently, the captured images may not fully reflect human perception.

\noindent\textbf{Limited sensing and display range.}
The compensation image is dynamically updated based on the viewer's tracked position. When the viewer moves beyond the Kinect's sensing range, however, compensation cannot be maintained. Moreover, because the projection area is limited, the compensated content may extend beyond the display boundary during viewpoint changes, resulting in truncation.

\section{Conclusion}
\label{sec:conclusion}
In this paper, we present a view-aware full projector compensation framework implemented with a single projector, a depth camera, and an RGB camera. To the best of our knowledge, this is the first approach to enable view-aware full projector compensation for a moving viewer in a naked-eye setting. To enable view-aware tracking, we introduce a metric co-calibration method that resolves the SfM scale ambiguity and registers the oppositely facing tracking camera. We further develop a non-iterative compensation framework that exploits Gaussian-based ProCams for novel-view rendering and geometric correction and photometric networks for single-pass compensation, resulting in VAComp-SD and VAComp-SI. Across 8 experimental cases, each evaluated at 78 novel viewpoints, both variants achieve competitive compensation quality. VAComp-SD supports real-time updates, while VAComp-SI eliminates per-setup photometric-network training while maintaining interactive-rate performance, demonstrating the practicality of VAComp for view-dependent projection mapping.

\bibliographystyle{abbrv-doi}
\bibliography{references}
\supptitle{VAComp: View-Aware Full Projector Compensation}
\makeMySupplementaryTitle
\appendix  
\setcounter{page}{1}
\section{Additional Qualitative Comparisons}
In the supplementary material, we provide additional qualitative comparisons
for the methods listed in \cref{tab:comparison_methods}. The
corresponding results are shown in
\cref{fig:supp_comparison_2,fig:supp_comparison_1}.
\section{Additional Comparisons of Pose Assumptions}
We provide additional comparisons between the speckle-calibrated pose and
the center-facing pose assumption in \cref{fig:supp_pose_assumptions}.
\section{Additional Computational Performance}
Because compensation is deployed on the server, we separately measure the
client- and server-side memory requirements during dynamic compensation.
\cref{tab:resource_usage} reports both average and peak RAM and GRAM usage,
together with the GRAM required for GS and photometric compensation network training.

\medskip
\noindent\begin{minipage}{\linewidth}
    \centering
    \includegraphics[width=\linewidth]{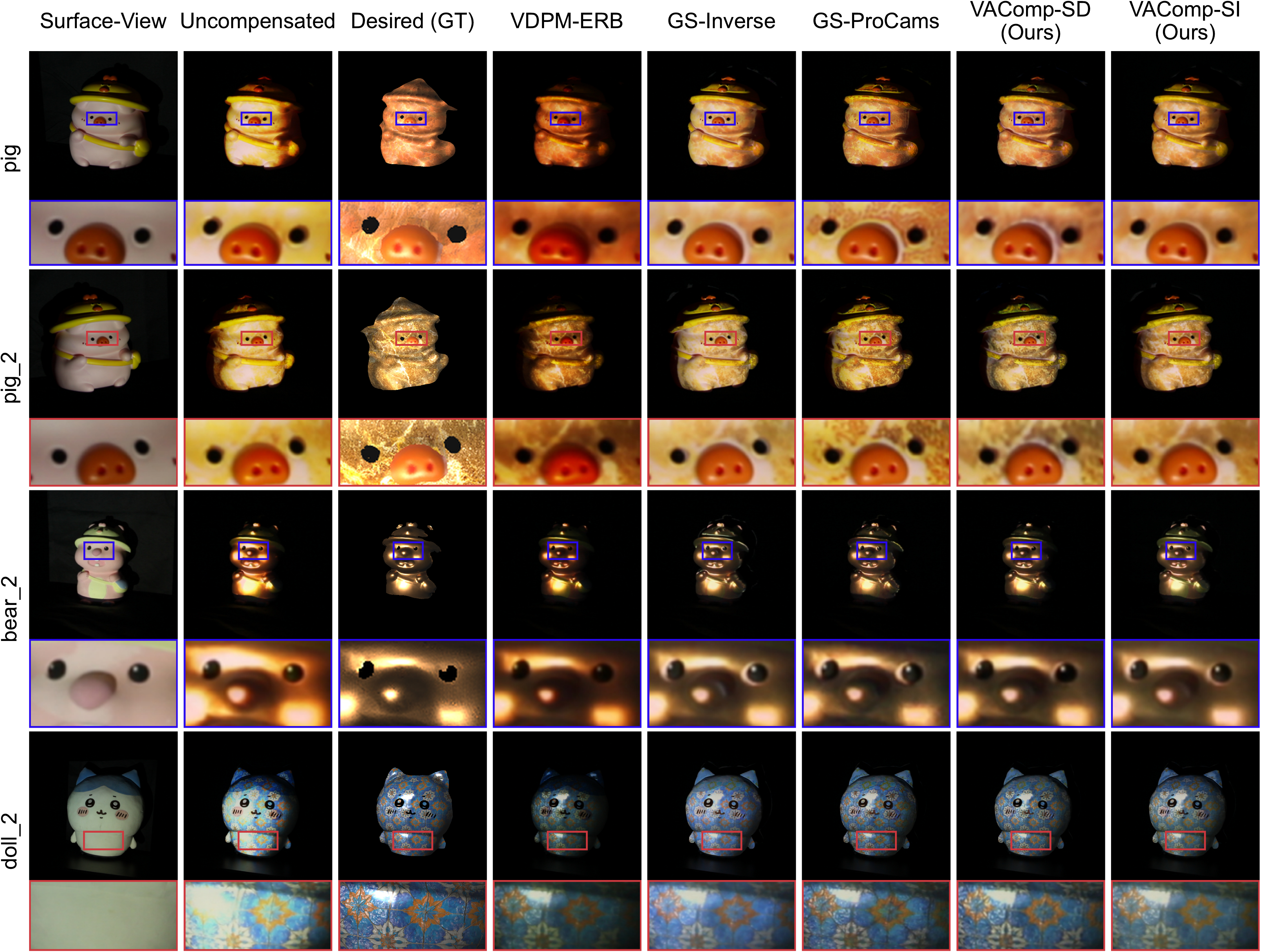}
    \captionof{figure}{\textbf{Visual comparisons of projector compensation for novel projections.} Each row represents a different surface, with the label at the beginning of the row indicating the name of the corresponding case. The first two columns show the camera-captured projection surface and its uncompensated images, while the remaining columns present the viewer virtual object image together with the camera-captured compensated projections produced by different methods. The blue and red boxes identify the regions shown in the corresponding detail rows. For clearer visualization, the images in the first three rows have been cropped and enlarged.}
    \label{fig:supp_comparison_1}
\end{minipage}

\medskip
\noindent\begin{minipage}{\linewidth}
    \centering
    \includegraphics[width=\linewidth]{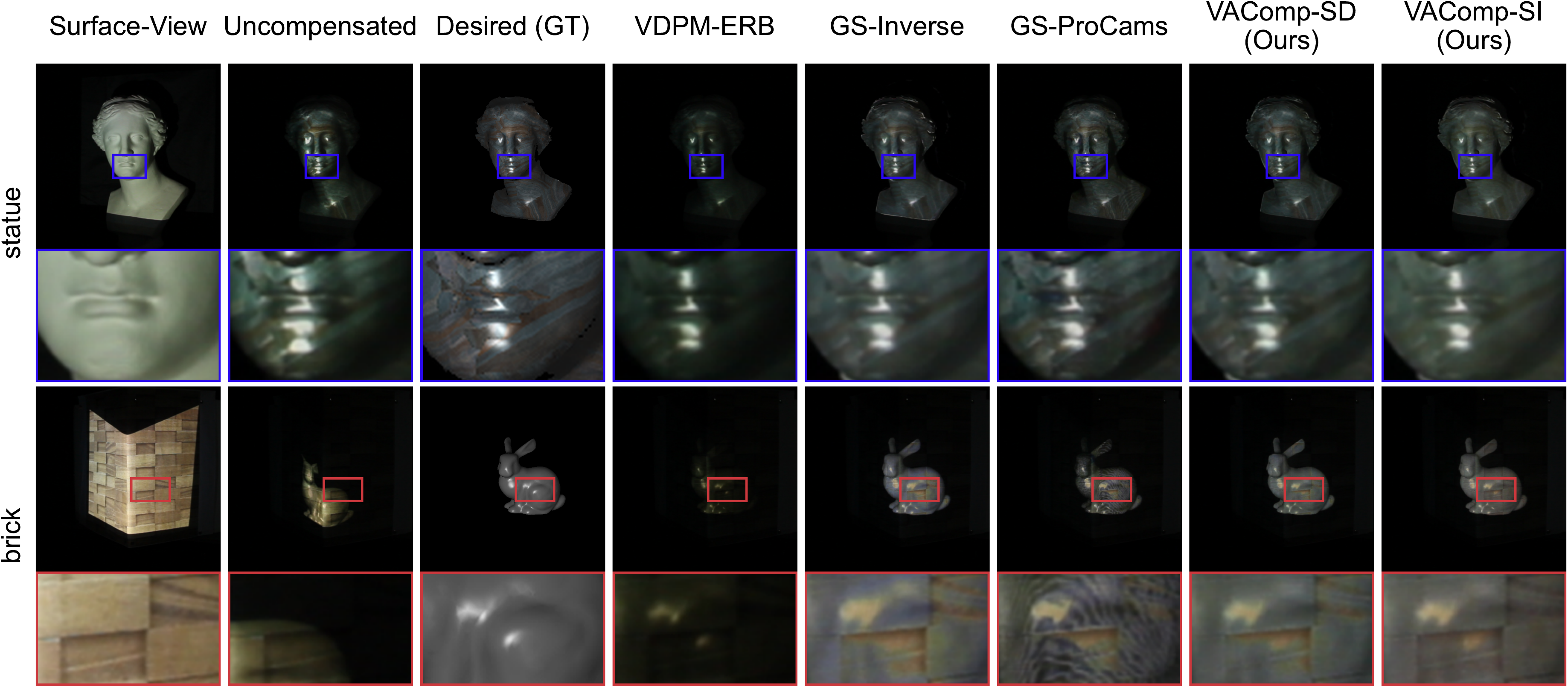}
    \captionof{figure}{\textbf{Visual comparisons of projector compensation for novel projections.} Each row represents a different surface, with the label at the beginning of the row indicating the name of the corresponding case. The first two columns show the camera-captured projection surfaces and their uncompensated images, while the remaining columns present the viewer virtual object images together with the camera-captured compensated projections produced by different methods. For clearer visualization, the images in both rows have been cropped and enlarged.}
    \label{fig:supp_comparison_2}
\end{minipage}


\medskip
\noindent\begin{minipage}{\linewidth}
    \centering
    \begin{minipage}[t]{0.47\linewidth}
        \centering
        \includegraphics[width=\linewidth]{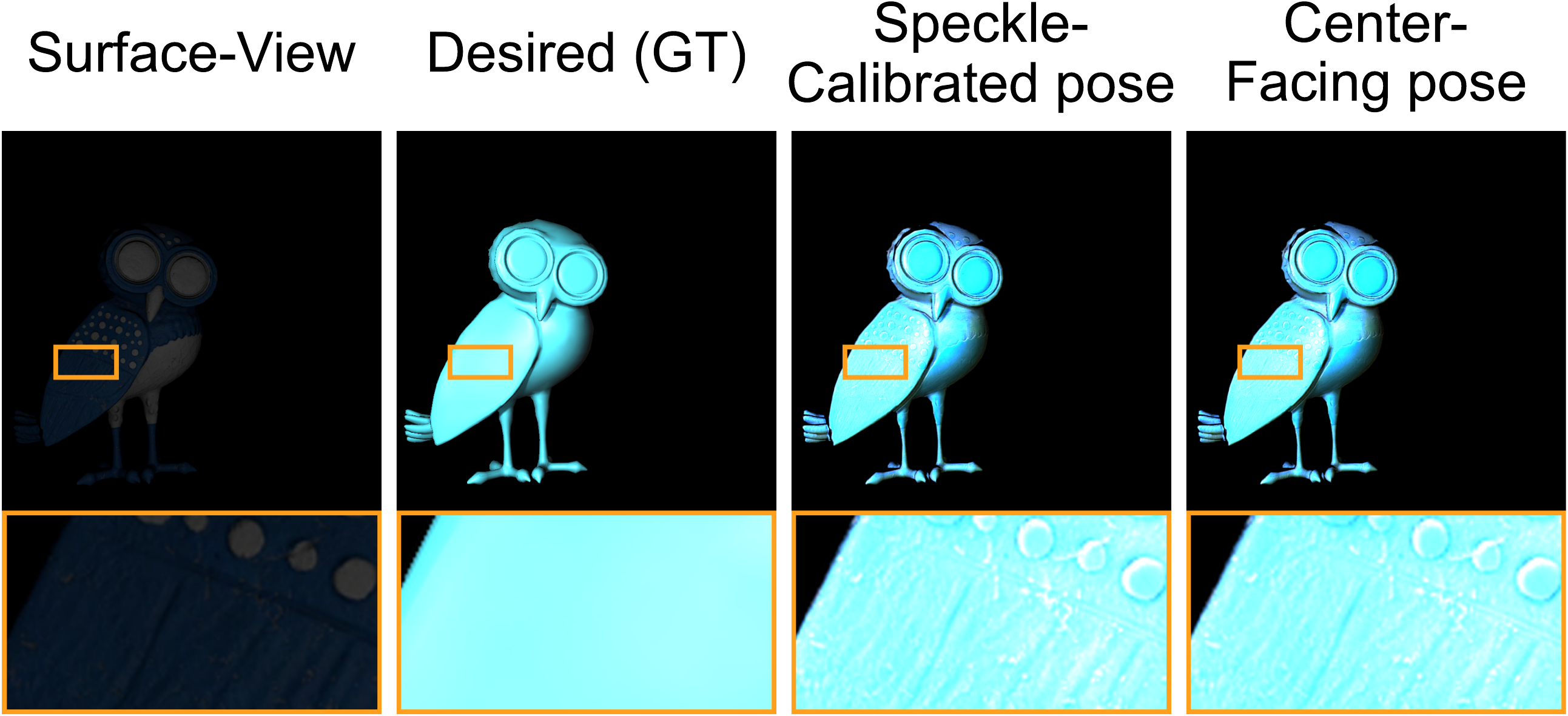}
    \end{minipage}\hfill
    \begin{minipage}[t]{0.47\linewidth}
        \centering
        \includegraphics[width=\linewidth]{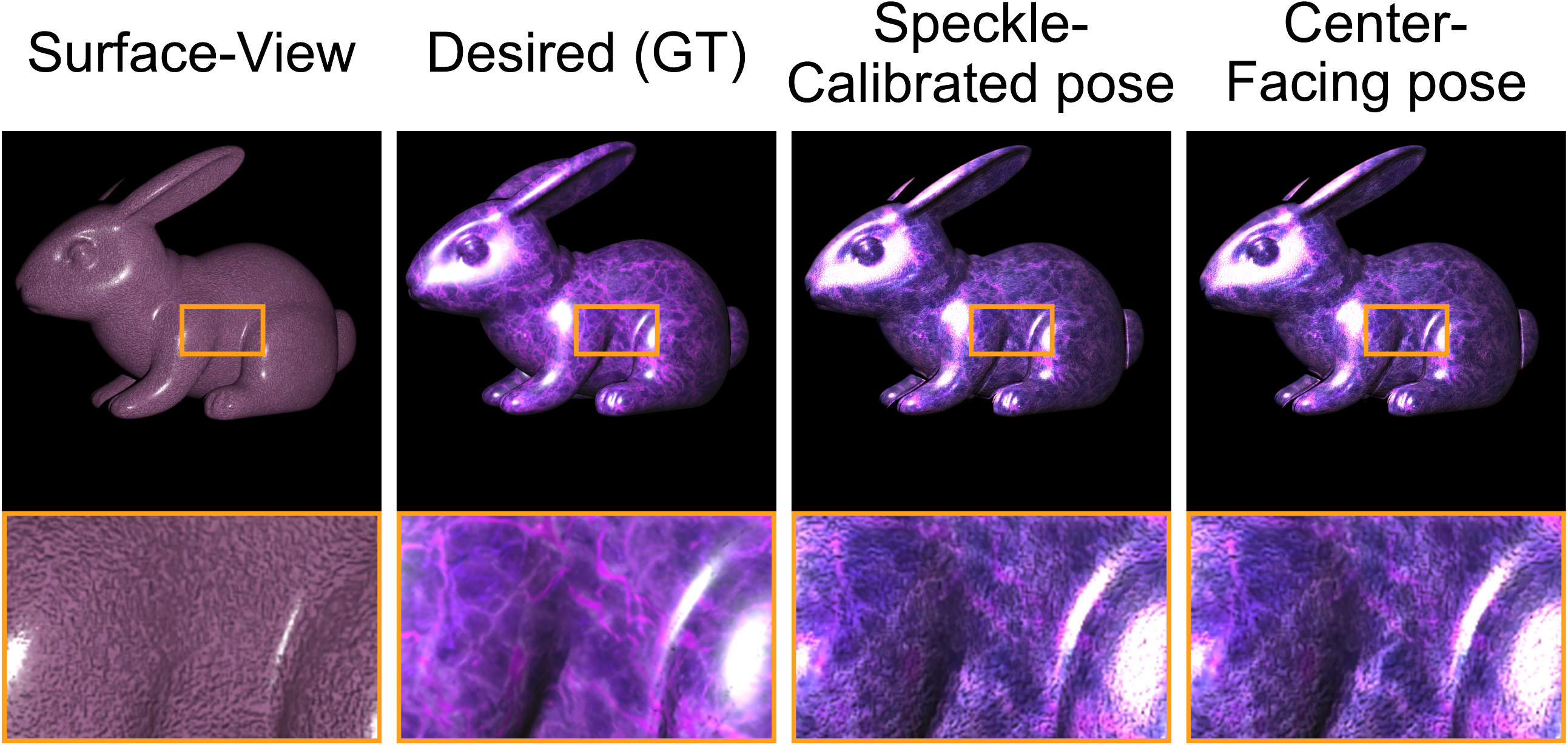}
    \end{minipage}

    \vspace{2pt}

    \begin{minipage}[t]{0.47\linewidth}
        \centering
        \includegraphics[width=\linewidth]{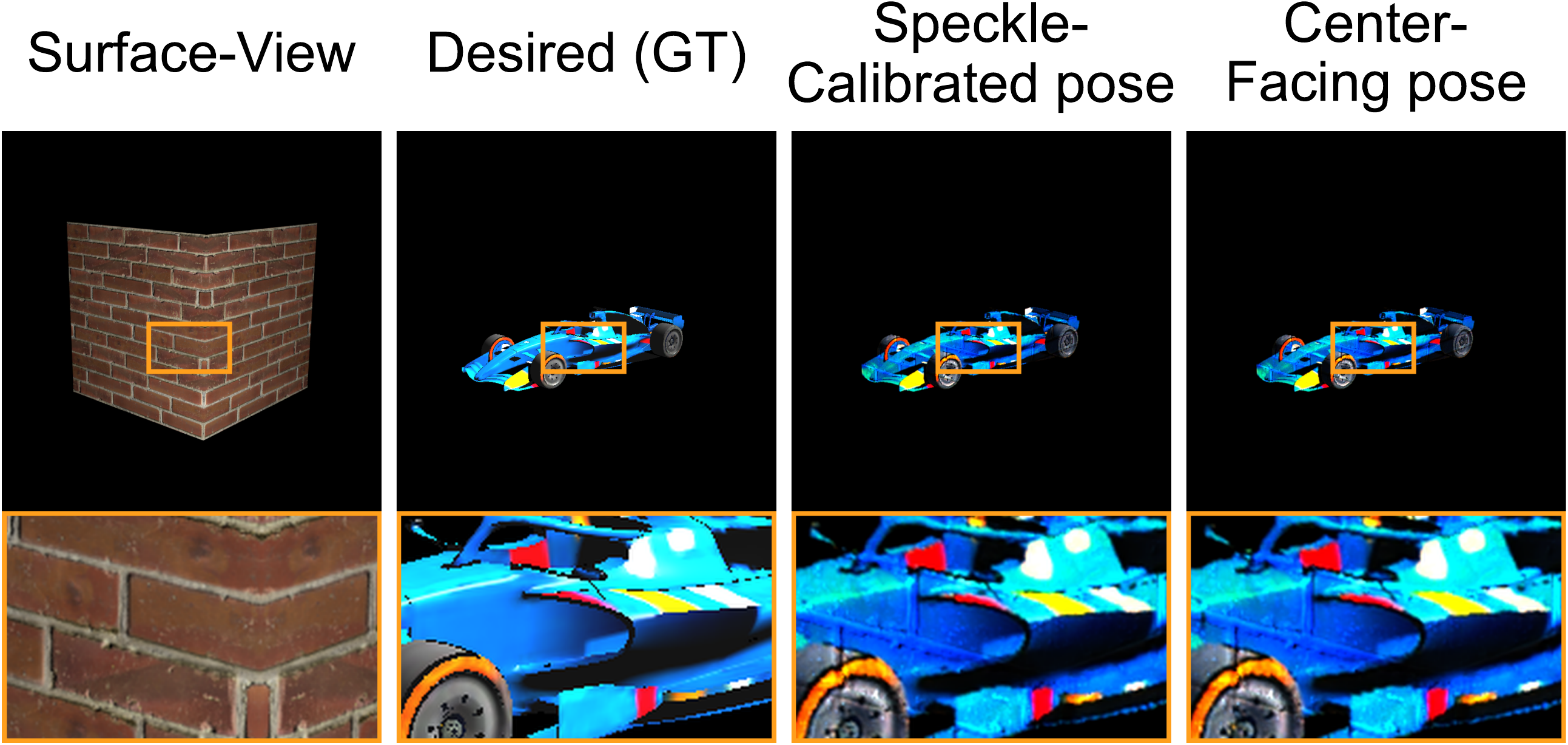}
    \end{minipage}\hfill
    \begin{minipage}[t]{0.47\linewidth}
        \centering
        \includegraphics[width=\linewidth]{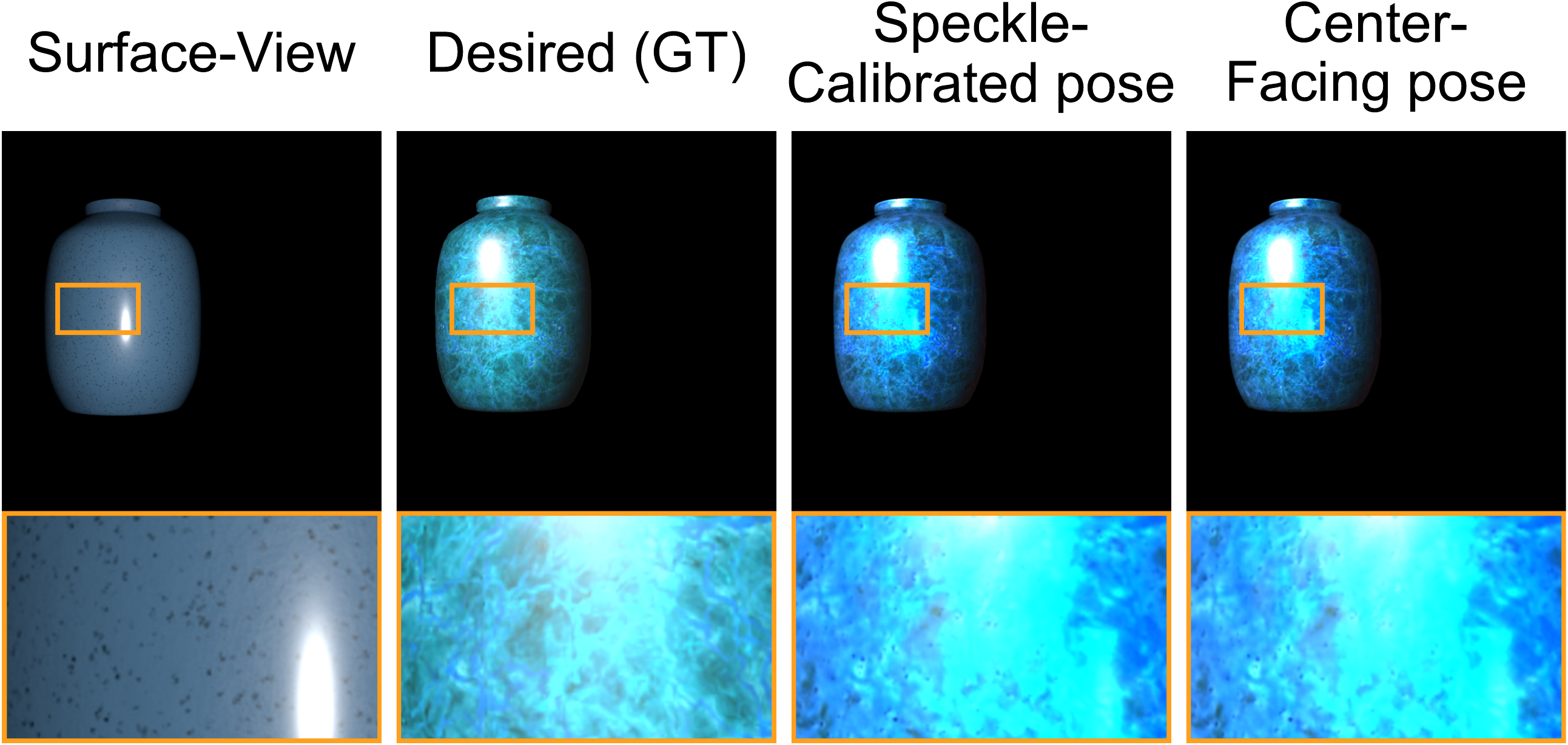}
    \end{minipage}
    \captionof{figure}{\textbf{Comparison of pose assumptions for projector compensation using VAComp-SI.} From left to right, the columns show the simulated projection surface, the viewer virtual object image, and the simulated compensated projections generated by VAComp-SI using the speckle-calibrated pose and the center-facing pose assumption, respectively. The bottom row shows magnified views of the regions highlighted by the orange boxes.}
    \label{fig:supp_pose_assumptions}
\end{minipage}


\clearpage


\medskip
\noindent\begin{minipage}{\textwidth}
\centering
\renewcommand{\arraystretch}{0.90}

\captionof{table}{
\textbf{Memory usage during dynamic compensation and training.}
(a) reports the average and peak RAM and GRAM on the client and server during dynamic compensation.
(b) reports the average and peak GRAM used to train the GS model and photometric compensation network (PCN).
All values are in GiB, and lower is better.
``-'' indicates that the corresponding measurement is not applicable.
For PCN training, zero indicates that no per-setup PCN training is required.
The best and second-best results are highlighted in \textbf{bold} and
\underline{underlined}, respectively, based on the unrounded values.
}
\label{tab:resource_usage}

\vspace{0pt}
\textbf{(a) Dynamic compensation}

\setlength{\tabcolsep}{3.5pt}
\begin{tabular*}{\textwidth}{@{\extracolsep{\fill}}lrrrrrrrr@{}}
\toprule
\textbf{Method}
& \multicolumn{4}{c}{\textbf{Client}}
& \multicolumn{4}{c}{\textbf{Server}} \\
\cmidrule(lr){2-5}
\cmidrule(lr){6-9}

& \multicolumn{2}{c}{\textbf{RAM}$\downarrow$}
& \multicolumn{2}{c}{\textbf{GRAM}$\downarrow$}
& \multicolumn{2}{c}{\textbf{RAM}$\downarrow$}
& \multicolumn{2}{c}{\textbf{GRAM}$\downarrow$} \\
\cmidrule(lr){2-3}
\cmidrule(lr){4-5}
\cmidrule(lr){6-7}
\cmidrule(lr){8-9}

& \textbf{Avg.} & \textbf{Peak}
& \textbf{Avg.} & \textbf{Peak}
& \textbf{Avg.} & \textbf{Peak}
& \textbf{Avg.} & \textbf{Peak} \\
\midrule

VDPM-ERB~\cite{KhuslenBattulga2021}
& \underline{13.81~GiB}
& \underline{14.18~GiB}
& 1.21~GiB
& 1.25~GiB
& \underline{3.03~GiB}
& \underline{3.08~GiB}
& \underline{1.04~GiB}
& \underline{1.08~GiB} \\

GS-Inverse
& \textbf{13.77~GiB}
& \textbf{14.03~GiB}
& 1.21~GiB
& \underline{1.25~GiB}
& \textbf{3.00~GiB}
& \textbf{3.06~GiB}
& \textbf{1.03~GiB}
& \textbf{1.08~GiB} \\

GS-ProCams~\cite{deng2025}
& 14.24~GiB
& 14.97~GiB
& \underline{1.21~GiB}
& 1.25~GiB
& 3.32~GiB
& 3.41~GiB
& 1.05~GiB
& 1.09~GiB \\

VAComp-SD (Ours)
& 14.02~GiB
& 14.29~GiB
& \textbf{1.20~GiB}
& \textbf{1.25~GiB}
& 3.17~GiB
& 3.23~GiB
& 1.08~GiB
& 1.11~GiB \\

VAComp-SI (Ours)
& 13.99~GiB
& 14.40~GiB
& 1.21~GiB
& 1.25~GiB
& 3.43~GiB
& 3.62~GiB
& 1.63~GiB
& 1.86~GiB \\

\midrule
Uncompensated
& - & - & - & -
& - & - & - & - \\
\bottomrule
\end{tabular*}

\vspace{2pt}
\textbf{(b) Training}

\setlength{\tabcolsep}{6pt}
\begin{tabular*}{\textwidth}{@{\extracolsep{\fill}}lrrrr@{}}
\toprule
\textbf{Method}
& \multicolumn{2}{c}{\textbf{GS GRAM}$\downarrow$}
& \multicolumn{2}{c}{\textbf{PCN GRAM}$\downarrow$} \\
\cmidrule(lr){2-3}
\cmidrule(lr){4-5}

& \textbf{Avg.}
& \textbf{Peak}
& \textbf{Avg.}
& \textbf{Peak} \\
\midrule

VDPM-ERB~\cite{KhuslenBattulga2021}
& \underline{0.76~GiB}
& 1.29~GiB
& -
& - \\

GS-Inverse
& 0.76~GiB
& \underline{1.28~GiB}
& -
& - \\

GS-ProCams~\cite{deng2025}
& \textbf{0.53~GiB}
& \textbf{1.07~GiB}
& -
& - \\

VAComp-SD (Ours)
& \underline{0.76~GiB}
& 1.29~GiB
& \underline{0.17~GiB}
& \underline{1.27~GiB} \\

VAComp-SI (Ours)
& \underline{0.76~GiB}
& 1.29~GiB
& \textbf{0.00~GiB}
& \textbf{0.00~GiB} \\

\midrule
Uncompensated
& - & - & - & - \\
\bottomrule
\end{tabular*}

\end{minipage}

\end{document}